\documentclass{llncs}
\usepackage[utf8]{inputenc}

\usepackage{amsmath,amssymb,amsfonts}
\usepackage{algorithmic}
\usepackage{graphicx}
\usepackage{textcomp}
\usepackage{xcolor}
\usepackage{colortbl}

\usepackage{siunitx}
\usepackage{microtype}                 % use micro-typography (slightly more compact, better to read)
\usepackage{tabu}                      % only used for the table example
\usepackage{booktabs}    
\usepackage{todonotes}

\usepackage{hyperref}
\usepackage{ dsfont }
\usepackage{xspace}
\usepackage{multirow}
\usepackage{subcaption}
\usepackage{etoolbox}
\usepackage{rotating}
\usepackage{makecell}
\usepackage{stmaryrd}
\usepackage[sort]{cite}

\title{Deep Neural Network for DrawiNg Networks, \modelname}
\titlerunning{\modelname}  % abbreviated title (for running head)
\author{Loann~Giovannangeli\orcidID{0000-0002-9395-6495} \and
Frederic~Lalanne\orcidID{0000-0001-9108-0955} \and
David~Auber\orcidID{0000-0002-1114-8612} \and
Romain~Giot\orcidID{0000-0002-0638-7504} \and
Romain~Bourqui\orcidID{0000-0002-1847-2589}
}

\authorrunning{Loann~Giovannangeli et al.} % abbreviated author list (for running head)
\institute{LaBRI, UMR CNRS 5800, University Bordeaux, 33405 Talence, France\\
\email{\{firstname\}.\{lastname\}@u-bordeaux.fr}
}

\newcommand{\ie}{\textit{i.e.},\xspace}
\newcommand{\eg}{\textit{e.g.},\xspace}
\newcommand{\etal}{\textit{et al.}\xspace}

\newcommand{\Nmax}{$N_{max}$\xspace}
\newcommand{\CardV}{$|$V$|$\xspace}
\newcommand{\CardE}{$|$E$|$\xspace}
\newcommand{\modelname}{\textit{(DNN)\textsuperscript{2}}\xspace}
\newcommand{\modelnameStar}{\textit{(DNN)\textsuperscript{2*}}\xspace}
\newcommand{\sgdgd}{\textit{S\_GD\textsuperscript{2}}\xspace}
\definecolor{mgrey}{rgb}{0.835,0.91,0.831}

\begin{document}

\maketitle              % typeset the title of the contribution

\begin{abstract}
By leveraging recent progress of stochastic gradient descent methods, several works have shown that graphs could be efficiently laid out through the optimization of a tailored objective function. 
In the meantime, Deep Learning (DL) techniques achieved great performances in many applications. We demonstrate that it is possible to use DL techniques to learn a graph-to-layout sequence of operations thanks to a graph-related objective function.
In this paper, we present a novel graph drawing framework called \modelname: Deep Neural Network for DrawiNg Networks. Our method uses Graph Convolution Networks to learn a model. Learning is achieved by optimizing a graph topology related loss function that evaluates \modelname generated layouts during training.
Once trained, the \modelname model is able to quickly lay any input graph out. We experiment \modelname and statistically compare it to optimization-based and regular graph layout algorithms. The results show that \modelname performs well and are encouraging as the Deep Learning approach to Graph Drawing is novel and many leads for future works are identified.
\end{abstract}

\section{Introduction}
\label{sec:introduction}
Optimization-based (OPT) and Deep Learning (DL) methods are gaining increasing interest in the information visualization field~\cite{ml4vis,ai4vis}. From the very design of visualizations to their evaluations, such techniques have shown to perform well and present benefits over standard methods. These advances motivated the exploration of these techniques adaptation to graph drawing.
Some studies \cite{gd2,zheng2018sgdgd,kruiger2017tsnet} used OPT approaches to optimize an objective function for a single graph with Stochastic Gradient Descent (SGD) and obtained good results; Zheng \etal~\cite{zheng2018sgdgd} even outperformed some state-of-the-art layout algorithms. 
On the other hand, if DL techniques have been applied on graph and graph drawing related problems  (\eg evaluate aesthetic metrics)~\cite{giovannangeli2020methodo,haleem2019evaluating,kwanliuDL4G}, to the best of our knowledge, only one study made use of this technique to \textit{draw} graphs, \textit{DeepDrawing}~\cite{wang2019deepdrawing}. 
Their framework leverages DL techniques to learn a model to reproduce layouts (\textit{i.e.}, ground truths) by optimizing a Procrustes-based cost function that compares the produced layout to the ground truth one. 
A major flaw of optimizing such a cost function by opposition to a graph topology related function is that the model is trained to optimize a similarity to a ground truth graph layout (that can be suboptimal) rather than the emphasis of the topology. 
This paper presents Deep Neural Networks for DrawiNg Networks, \modelname, a graph layout framework relying on \textit{unsupervised} Deep Learning. It proposes to adapt well-proven Convolutional Neural Network architecture to graph context using Graph Convolutions~\cite{cnn4graph,kipf2016gcn}. To the best of our knowledge, it is the first Deep Neural Network (DNN) architecture trained to lay generic graphs out by optimizing a graph-drawing related cost function. 
We propose an experimentation of \modelname where we use \textit{ResNet}~\cite{he2016resnet} architecture as a basis to optimize the Kruiger \etal~\cite{kruiger2017tsnet} adaptation of the \textit{Kullback-Leibler} divergence. In DL, as a model performance and its capability to generalize to unseen data are often incompatible, we also study the benefits of pre-training \modelname. 
Finally, we statistically compare \modelname with state-of-the-art methods on aesthetic metrics and find that it competes with them. By efficiently learning a bounded sequence of operations that lays generic graphs out, \modelname experimentation suggests that graph drawing can be modeled as a mathematical function.

The remainder of the paper is organized as follows. \autoref{sec:related_works} presents related works on OPT and DL methods in graphs context. \autoref{sec:framework} introduces \modelname and its key concepts while \autoref{sec:benchmark} presents the results of its experimental evaluation. \autoref{sec:discussion} discusses the visual aspect of \modelname layouts and its limitations. Conclusions and leads for future works are presented in \autoref{sec:conclusion}.

\section{Related Works}
\label{sec:related_works}
First, we define the conventional notations used in this paper. Let $G(V,E)$ be a graph: $V$ is its set of nodes $\{v_i\}, i \in [1, N], N = |V|$
and $E \subseteq V \times V$ its set of edges. 
Graphs are considered simple and connected.
Let nodes positions be encoded in a vector $X \in \mathds{R}^{N\times2}$ where $X_i$ is the 2D position of node $v_i$, $\left||X_i - X_j\right||$ relates to the Euclidean distance between points $X_i$ and $X_j$.

Optimization-based (OPT) and Deep Learning (DL) techniques applications to Graph Drawing are gaining popularity and have been applied to a variety of graph and graph drawing related problems. For instance, Kwon \etal~\cite{kwanliuML4G} used Machine Learning techniques to approximate a graph layout and its aesthetic metrics at the same time. Haleem \etal~\cite{haleem2019evaluating} also proposed to predict aesthetic metrics using a DL model. Several studies ~\cite{deepwalk,Line,grover2016node2vec} used OPT to compute a feature vector embedding of a graph nodes. Kwon and Ma~\cite{kwanliuDL4G} proposed a Deep encoder-decoder to learn smooth transitions between different layouts of a graph.

Recently, OPT and DL techniques were proposed to lay graphs out and did compete with state-of-the-art layout algorithms. 
Kruiger \etal~\cite{kruiger2017tsnet} proposed to optimize the Kullback-Leibler divergence by gradient descent. Kullback-Leibler divergence is a measure of dissimilarity between two probabilities distribution $P$ and $Q$ which was used to visualize data~\cite{sne,tsne} and is defined as: $D_{KL} = \sum\limits_i P(i) \log\frac{P(i)}{Q(i)}$. The proposed optimization framework, $tsNET$, showed to perform well, although its execution time is extremely high (\ie several seconds for graphs with $N < 100$). The authors proposed an improved variant of their method for which nodes positions are initialized with PivotMDS rather than randomly. This variant showed to be more efficient in terms of aesthetic metrics and converged faster on larger graphs. 
\sgdgd~\cite{zheng2018sgdgd} relies on the optimization of \textit{stress} by stochastic gradient descent (SGD). Stress is modeled by a set of constraints between nodes that are relaxed by iteratively moving pairs of nodes. 
$GD^2$~\cite{gd2} also leveraged SGD to optimize a set of aesthetic metrics whose combination can be tuned by associating a weight to each metric. 

On the other hand, GraphTSNE~\cite{graphtsne} learned a shallow Neural Network made of Graph Convolutions to predict a graph layout. The key idea of their work is to train a model for each graph to draw, the train dataset being the graph nodes themselves. Even if their model cannot be described as \textit{deep}, their work confirms that a t-SNE based loss can be optimized by Graph Convolutions networks.
\textit{DeepDrawing}~\cite{wang2019deepdrawing} is the first method to train a DNN to compute graph layouts. It aims to mimic a \textit{target} algorithm given as ground truth and can be seen as a fast approximation of its \textit{target}. This was also studied by Espadoto \etal~\cite{espadoto2020deep} and both studies raised several limitations to this approach. First, it requires to run the \textit{target} algorithm thousands of times to generate labeled training data. Due to model convergence issue, the labeled data generation should be manually supervised and the model cannot reproduce results of a non-deterministic algorithm either. Second, as the model learns to mimic an algorithm, it cannot produce better results than its \textit{target} baseline and it also learns its defects. 
Finally, as the function optimized by the model is not related only to its input data, it does not learn features from its input but rather from its combination input--target algorithm. Hence, it is unclear how well it can generalize to unseen data for which no \textit{target} result was ever provided. 
As opposed to DeepDrawing, \modelname training is \textit{unsupervised} (\ie no groundtruth layout is provided) and generated graph layouts are evaluated according to a graph topology related cost function based on t-SNE.

\section{\modelname Framework Design}
\label{sec:framework}
\subsection{\modelname Architecture}
\label{sec:architecture}
\subsubsection{A ResNet-like Basis.}

\begin{figure}[!tb]
\centering
\includegraphics[width=\columnwidth]{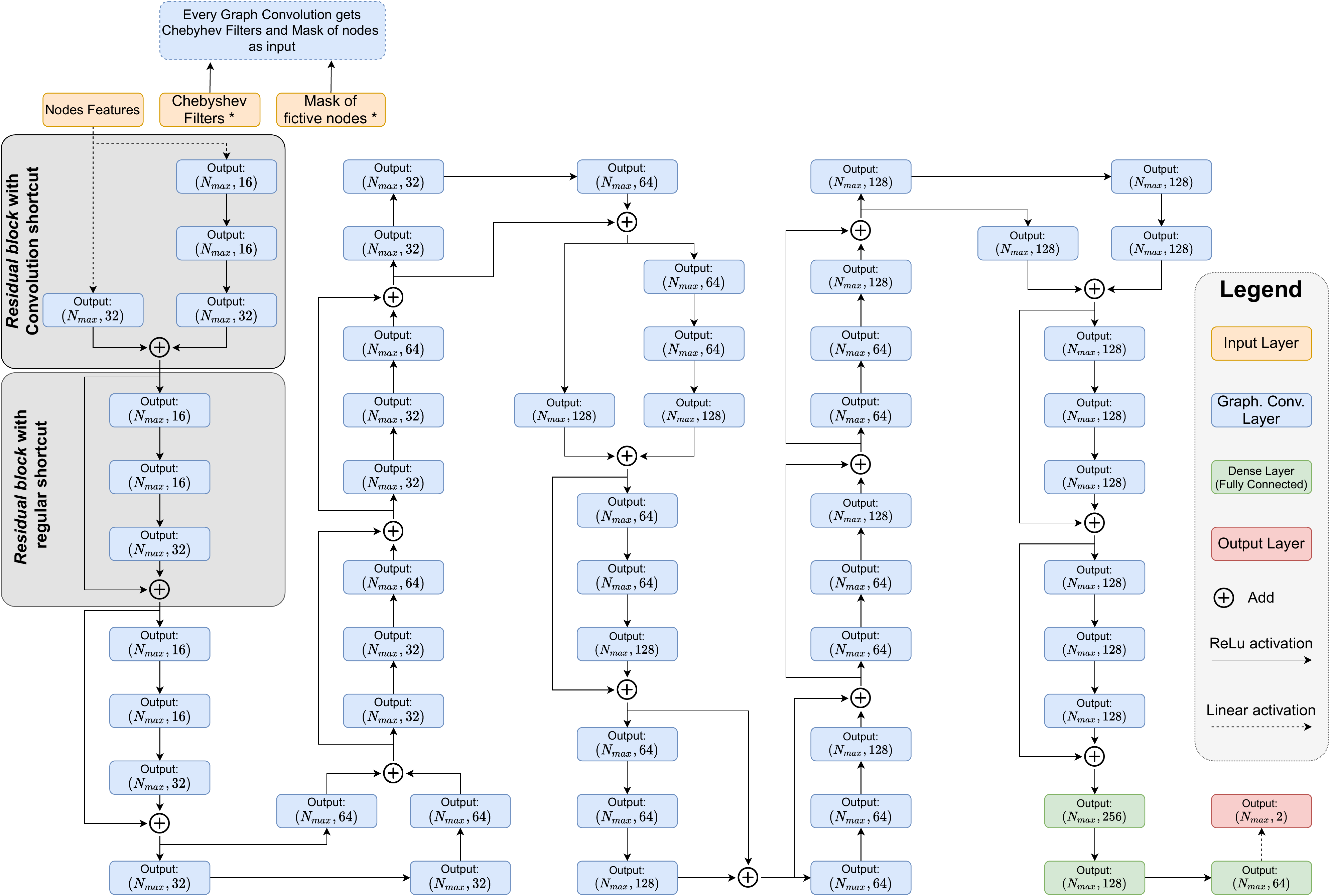}
\caption{\modelname architecture based on ResNet50~\cite{he2016resnet}. Some details have been omitted: (i) Chebyshev filters input is provided up to order 4 to all Graph Convolution layers except the 9 last layers (up to order 2); (ii) features vectors (\ie convolutions outputs) are normalized after every convolution; (iii) normalized features vectors are applied a mask encoding the \textit{real-fictive} nodes information; and (iv) only the first two residual blocks are emphasized out of the 16 blocks.}
\label{fig:archi}
\end{figure}

The design of \modelname architecture leverages Convolutional Neural Networks (CNNs) by adapting them to a graph context with Graph Convolutions~\cite{cnn4graph,kipf2016gcn}. The architecture reproduces \textit{ResNet}~\cite{he2016resnet}, a CNN designed to classify images and reaching a high accuracy on the ImageNet challenge~\cite{deng2009imagenet}. It is composed of \textit{residual blocks} that contain \textit{shortcuts} which enable the model to work on several levels of abstraction. It is made of 52 spectral \textit{Graph Convolutions} (see \autoref{sec:graph_conv}) organized in 16 residual blocks as in the ResNet architecture (see \autoref{fig:archi}). In addition, three node-wise fully connected layers with shared weights are added after the last convolution, the final layer being the model output. To handle graphs of varying sizes, the model inputs are fixed to an arbitrary size \Nmax and are padded with \textit{fictive nodes} to fit this size. After each residual block, its resulting features tensor is multiplied with a mask of real-fictive nodes $Mask \in \mathds{1}^{N_{max}}$ where $Mask_i = 0$ if $v_i$ is a fictive node, $1$ otherwise. 
Padding the model inputs to match the expected shape could create a bias during the training: if fictive nodes are always padded at the same position in the tensors, some trainable weights will mostly see irrelevant features of \textit{fictive nodes} and be underfitted. To avoid this bias, the padded model inputs are randomly permuted.

\subsubsection{Spectral Graph Convolutions.}
\label{sec:graph_conv}

Abbreviated Graph Convolutions, they were defined by Kipf and Welling~\cite{kipf2016gcn} to operate on a graph \textit{signal} encoded as a features vector for every node. The convolution kernel size $K$ is defined to convolve a node with its $K$-hop neighborhood. The graph topology is provided through the graph spectrum (\ie eigendecomposition of the normalized Laplacian matrix)~\cite{cohen2018approximating}, approximated with Chebyshev polynomials~\cite{hammond2011cheby}. 
Graph Convolutions are formally defined as a function of a signal $x$:

\begin{figure}[!tb]
\centering
\includegraphics[width=\columnwidth]{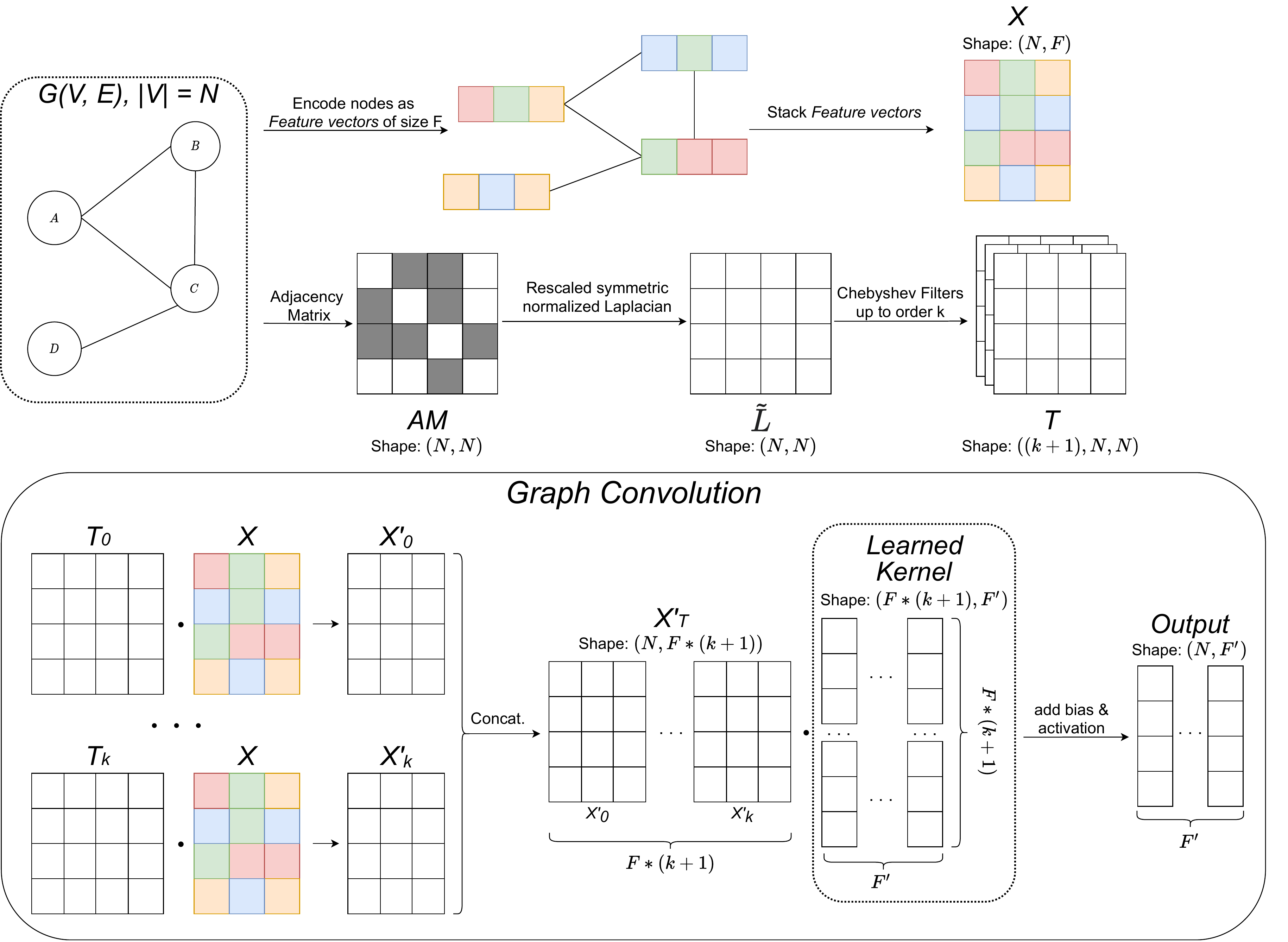}
\caption{Graph Convolutional layer diagram. It takes two tensors as input: (i) a feature vector to convolve ($X$); and (ii) a data structure that encodes the graph topology ($T$). $X$ can refer to a node features vector at any step of the training.}
\label{fig:graph_conv}
\end{figure}

\begin{equation}
\label{eq:graph_conv0}
g_\theta \star x = Ug_\theta U^T x
\end{equation}
where $U$ is the matrix of eigenvectors of the symmetric normalized Laplacian matrix $L$ so that $L = U\Lambda U^T$ where $\Lambda$ are the eigenvalues of $L$. As the evaluation of \autoref{eq:graph_conv0} and the eigendecomposition of $L$ are expensive, the operation can be approximated~\cite{hammond2011cheby} using the Chebyshev polynomials $T_k(x)$ up to order $K$:

\begin{equation}
g_{\theta'} \star x \approx \sum_{k=0}^{K} \theta'_k T_k\left(\tilde{L}\right) x
\end{equation}
where $\tilde{L}$ is the rescaled symmetric normalized Laplacian $\tilde{L} = \frac{2}{\lambda_{max}}L-I_N$, $\lambda_{max}$ being the highest eigenvalue in $\Lambda$ and $I_N$ the identity matrix of size $N$. $\theta' \in \mathds{R}^K$ is a vector of Chebyshev coefficients and $T_k(x)$ is the Chebyshev polynomial defined as $T_0(x) = 1$, $T_1(x) = x$ and $T_k(x) = 2x T_{k-1}(x) - T_{k-2}(x), \forall k\ge2$ and that costs $\mathcal O(K|E|)$ to be computed up to order $K$~\cite{hammond2011cheby}.
The Graph Convolution computation in this paper (illustrated in \autoref{fig:graph_conv}) can be formally defined as:

\begin{equation}
Z = \bigparallel_{k=0}^K T_k\left(\tilde{L}\right) \cdot X \cdot \Theta
\end{equation}
where $X \in \mathds{R}^{N \times F}$ is the nodes features vectors (\ie graph signal) where each node has $F$ features, and $\Theta \in \mathds{R}^{(F*(K+1)) \times F'}$ is the learned graph convolution kernel where $F'$ is the size of the desired output feature vector for every node. The symbol $\bigparallel$ is used as a \textit{concatenate} operator on all the $T_k(\tilde{L}) \cdot X$ tensors. \\

Finally, \modelname is fed with three tensors: the graph signal (nodes feature vectors, defined later in \autoref{sec:nodes_features}), a mask of real-fictive nodes and the Chebyshev polynomials (also referred to as \textit{Chebyshev filters}). Its output is set to a \Nmax$ \times 2$ tensor of nodes positions in the plane. 
The time complexity of a forward pass in the model is $\mathcal O(N_{max})$ as this constant bounds the tensors size.

\subsection{Loss function}
\label{sec:loss}
Unlike \textit{DeepDrawing}~\cite{wang2019deepdrawing}, \modelname is trained to optimize a loss function that captures the graph layout quality based on its topology. As optimizing a function for a whole dataset is fundamentally different from optimizing it for specific graphs, the loss function should have already been used with standard and OPT methods to lay graphs out so that we can compare their performances. This mainly let us with two possible functions: \textit{stress} and \textit{Kullback-Leibler(KL) minimization} (see \autoref{sec:related_works}). If we believe both can be optimized by \modelname, we selected the \textit{KL minimization} from Kruiger \etal~\cite{kruiger2017tsnet} as it adapted better to the framework throughout experimentations. The loss is then defined as:

\begin{equation}
\label{eq:loss}
C = \lambda_{KL}C_{KL} + \frac{\lambda_c}{2N}\sum_i ||X_i||^2 - \frac{\lambda_r}{2N^2} \sum_{i,j \in V, i \neq j} \log\left(||X_i - X_j|| + \epsilon_r\right)
\end{equation}
where $C_{KL}$ is the main topology-related cost term based on the Kullback-Leibler divergence proposed by Kruiger \etal~\cite{kruiger2017tsnet}. The second and third terms are respectively a \textit{compression} that minimizes the scale of the drawing and a \textit{repulsion} that counter-balances the compression. ($\lambda_{KL},\lambda_{c},\lambda_{r}$) are weights used to tune the loss function during the optimization. $\epsilon_r = \frac{1}{20}$ is a regularization constant.

Kruiger \etal~\cite{kruiger2017tsnet} defined two \textit{stages} for their $tsNET$ algorithm. In the first stage, the three $\lambda$ factors are set to ($\lambda_{KL}=1,\lambda_{c}=1.2,\lambda_{r}=0$) while in the second stage, they are switched to ($1, 0.01, 0.6$). They also proposed a variant called $tsNET^*$ with two differences: nodes positions are initialized with PivotMDS~\cite{pivotmds} and the first stage lambda factors are ($1,0.1,0$). In this paper, \modelname extends both $tsNET$ variants and is compared to their implementation\footnote{\url{https://github.com/HanKruiger/tsNET}, consulted on February 2021}.

\subsection{Graph signal: Initial \textit{Nodes Features}}
\label{sec:nodes_features}

The graph signal is defined as a features vector for every node. Some methods already exist to extract a graph signal~\cite{grover2016node2vec,deepwalk,Line}. 
As standard layout algorithms achieved to lay graphs out only using their topology~\cite{zheng2018sgdgd,kamada1989algorithm,frick1994gem}, we assume it can be sufficient to feed the model with this information encoded through Chebyshev filters.
\textit{Nodes features} are then represented by a tensor $F \in \mathds{R}^{N\times2}$ with nodes \textit{id} to help the model differentiate them and a \textit{random metric} to reduce overfitting.

With this \textit{nodes features} tensor, it can be expected that adding meaningful features should help the model achieving better layouts. We experimented additional features by adding PivotMDS 2D positions such as in $tsNET^*$ variant~\cite{kruiger2017tsnet}, raising its size to $F \in \mathds{R}^{N\times4}$. This tensor is then transformed throughout the model successive Graph Convolution and Dense layers as presented in \autoref{fig:archi}.

As all the nodes features are not necessarily of the same order of magnitude, they are normalized to give them the same importance.

\section{Experimentation and Statistical Comparison}
\label{sec:benchmark}
%%%%%%%%%%%%%%%%%%%%%%%%%%%%%%%%%%%%%%%%%%%%%%%%%%%%%%%%%%%%%%%%%%%%%%%%%%%%%%%
%%%%%%%%%%%%%%%%%%%%%%%%%%%%%%%%%%% DATASETS %%%%%%%%%%%%%%%%%%%%%%%%%%%%%%%%%%%
%%%%%%%%%%%%%%%%%%%%%%%%%%%%%%%%%%%%%%%%%%%%%%%%%%%%%%%%%%%%%%%%%%%%%%%%%%%%%%%

\subsection{Datasets}
\label{sec:datasets}
Two datasets were considered for this experimentation (see \autoref{tab:graph_datasets}), both being split for Deep Learning validation purposes (\ie \textit{hold out} validation). In our terminology, \textit{train} and \textit{validation} sets are used during training to feed the model and evaluate it. \textit{Test} set is used to benchmark models on unseen data.

\subsubsection{Random Graphs.} Used to pretrain \modelname, its \textit{train} set was generated to sample 1000 random graphs for each graph size between 2 and \Nmax. 
It is noteworthy that by generating 1000 instances of each graph size, the model will see many isomorphic graphs (mainly of small size). It means the model could overfit on small graphs, but this kind of overfitting could be beneficial for it. Since graphs can be decomposed into subgraphs of smaller size, the model capability to layout a small graph $g$ can help it laying out a larger graph $G$ having $g$ as a subgraph. The \textit{validation} set was generated with 200 instances per graph size. 

\subsubsection{Rome Graphs.} Rome is a dataset of undirected graphs provided by the Graph Drawing symposium\footnote{\url{http://www.graphdrawing.org/data.html}, consulted on February 2021} made of \num{11534} graphs, \num{3} of them being excluded as they are disconnected. The set was randomly split as presented in \autoref{tab:graph_datasets} and the layout methods of this experiment will be evaluated on the Rome \textit{test} set.    

\begin{table}[!tb]
\centering
\resizebox{0.8\columnwidth}{!}{
\begin{tabular}{|c|c|c|c|c|c|c|}
\cline{2-7}
 \multicolumn{1}{c|}{} & \multicolumn{3}{c|}{\textbf{Graphs distribution}} & \multicolumn{3}{c|}{\textbf{Dataset size}} \\
\cline{2-7}
\multicolumn{1}{c|}{} & \textbf{\CardV} & \textbf{\CardE} & \textbf{Degree} & \textbf{Train} & \textbf{Validation} & \textbf{Test} \\
\hline
 Random Graphs & [2, 128] & [1, 6502] & [1, 118] & \num{127000} & \num{25400} & -- \\
 \hline
 Rome Graphs & [10, 107] & [9, 158] & [1, 13] & \num{8000} & \num{1600} & \num{1931} \\
\hline
\end{tabular}
}
\vspace{0.1cm}
\caption{Random and Rome graphs datasets properties.}
\label{tab:graph_datasets}
\end{table}

%%%%%%%%%%%%%%%%%%%%%%%%%%%%%%%%%%%%%%%%%%%%%%%%%%%%%%%%%%%%%%%%%%%%%%%%%%%%%%%
%%%%%%%%%%%%%%%%%%%%%%%%%%%%%%%%%%% TRAINING %%%%%%%%%%%%%%%%%%%%%%%%%%%%%%%%%%%
%%%%%%%%%%%%%%%%%%%%%%%%%%%%%%%%%%%%%%%%%%%%%%%%%%%%%%%%%%%%%%%%%%%%%%%%%%%%%%%

\subsection{Training}
\label{sec:training}

\begin{figure}[!tb]
\centering
\includegraphics[width=\columnwidth]{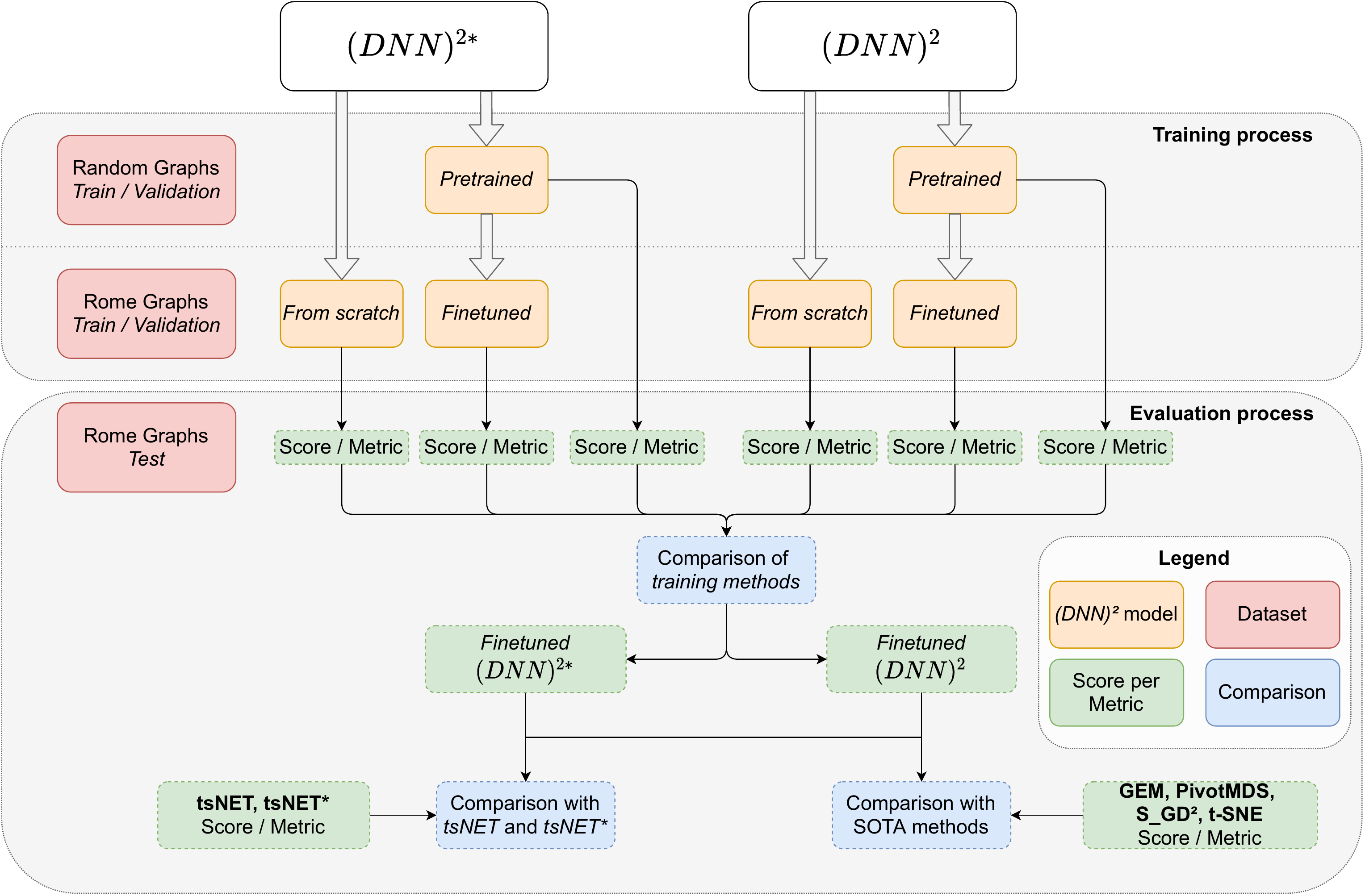}
\caption{\modelname training and evaluation pipeline. Six models are initially trained and compared. Then, the best selected models for the two loss variants are compared to $tsNET$, $tsNET^*$ and state-of-the-art layout methods.}
\label{fig:train_eval_pipe}
\end{figure}

In this experiment, \Nmax was set to \num{128} to be slightly larger than the biggest graph in the Rome dataset. 
Transformed features tensors sizes were defined experimentally and are described in \autoref{fig:archi}. Chebyshev filters were computed up to order 4 for all the Graph Convolution layers except the nine last ones which were only provided up to order 2. 
Therefore, more weight is given to direct neighborhood which minimizes overdraws that are critical for the drawing quality.

Since we want to compare our DL approach to the original $tsNET$ and $tsNET^*$ algorithms, an instance of \modelname is trained for each of them. We refer to these two variants as \modelname and \modelnameStar.
Similarly to $tsNET$, the models were trained in two stages. First, to optimize the loss $C$ (see \autoref{eq:loss}) with their respective $tsNET$ lambda weights (see \autoref{sec:loss}). Second, the optimizer is reset and models are trained to optimize $C$ with \textit{second stage} lambda weights.

\modelname variants were trained with three methods to be evaluated on Rome graphs: (i) \textit{pretraining} on Random graphs, (ii) \textit{finetuning} (after the \textit{pretraining}) on Rome graphs, (iii) training \textit{from scratch} on Rome graphs. The goal is to verify if pretraining the model on a large set of random graphs improves its performances, and whether training on a specific dataset leads to better results than on random graphs. There are six \modelname instances in total (see \autoref{fig:train_eval_pipe}).

The nodes features are rescaled in $[0; 1]$ based on the \textit{train} set. 
The random permutation of the model inputs (see \autoref{sec:architecture}) is fixed for the \textit{test} set graphs so that every model is evaluated on the same permuted graphs.

%%%%%%%%%%%%%%%%%%%%%%%%%%%%%%%%%%%%%%%%%%%%%%%%%%%%%%%%%%%%%%%%%%%%%%%%%%%%%%%
%%%%%%%%%%%%%%%%%%%%%%%%%%%%%%%%%%% METRICS %%%%%%%%%%%%%%%%%%%%%%%%%%%%%%%%%%%
%%%%%%%%%%%%%%%%%%%%%%%%%%%%%%%%%%%%%%%%%%%%%%%%%%%%%%%%%%%%%%%%%%%%%%%%%%%%%%%

\subsection{Metrics and comparison procedure}
\label{sec:eval_metrics}
Designing quality metrics for assessing a graph drawing quality that corroborates how well human subjects understand the drawing is a challenging question~\cite{purchase1995validating,purchase1997aesthetic,ware2002cognitive,purchase2002metrics}. We use a set of common metrics to assess \modelname efficiency and statistically compare it to state-of-the-art methods. Following the recommendations of Purchase~\cite{purchase2012experimental}, some metrics (marked with \textit{*}) were inverted so that all metrics can be read as \textit{lower is better} (see table~\autoref{tab:metrics}). In addition, we measured Execution times of each algorithm in milliseconds (\textit{ms}).
\begin{table}[!tb]
\centering
\begin{tabular}{|l|l|}
\hline
\multicolumn{1}{|c}{\bfseries Metric} & \multicolumn{1}{|c|}{\bfseries Reference} \\
\hline
\textit{Aspect ratio*} & As defined in~\cite{gd2}.\\
\textit{Angular resolution*} & As defined in~\cite{gd2}.\\
\textit{Edge crossings number} & Well-known aesthetic metric~\cite{gd2,purchase2002metrics}.\\
\textit{Cluster overlap} & Autocorrelation metric in~\cite{ambiguityvis} with MCL clustering~\cite{van2000mcl}. \\
\textit{Neighborhood preservation*} & As defined in~\cite{kruiger2017tsnet}.\\
\textit{Stress} & Well-known aesthetic metric~\cite{gd2}, normalized by $N$.\\
\hline
\end{tabular}
\vspace{0.1cm}
\caption{Quality metrics used in our benchmark and references to their definition. \textit{*} represents metrics inverted to allow a \textit{lower is better} reading for all of them.}
\label{tab:metrics}
\end{table}

In the next, the efficiency of different graph drawing techniques are statistically compared on the presented metrics. To assess which method performs significantly better, a Kruskal-Wallis test~\cite{kruskalwallis} first verifies whether the differences of performances between all the compared methods on a given metric are significant or not. If so, a post-hoc Conover test~\cite{conover} is applied to verify which pairs of methods are performing significantly different on that metric. For both tests, the acceptance threshold is set to $\alpha = 0.05$ and all Kruskal-Wallis tests passed.

%%%%%%%%%%%%%%%%%%%%%%%%%%%%%%%%%%%%%%%%%%%%%%%%%%%%%%%%%%%%%%%%%%%%%%%%%%%%%%%
%%%%%%%%%%%%%%%%%%%%%%%%%%%%%%%%%%% BENCH 1 %%%%%%%%%%%%%%%%%%%%%%%%%%%%%%%%%%%
%%%%%%%%%%%%%%%%%%%%%%%%%%%%%%%%%%%%%%%%%%%%%%%%%%%%%%%%%%%%%%%%%%%%%%%%%%%%%%%

\subsection{Training Methods Evaluation}
\label{sec:benchmark1}

\begin{figure}[!ht]
\centering
\resizebox{\columnwidth}{!}{
\begin{tabular}{ccc}
 \includegraphics[width=.31\columnwidth]{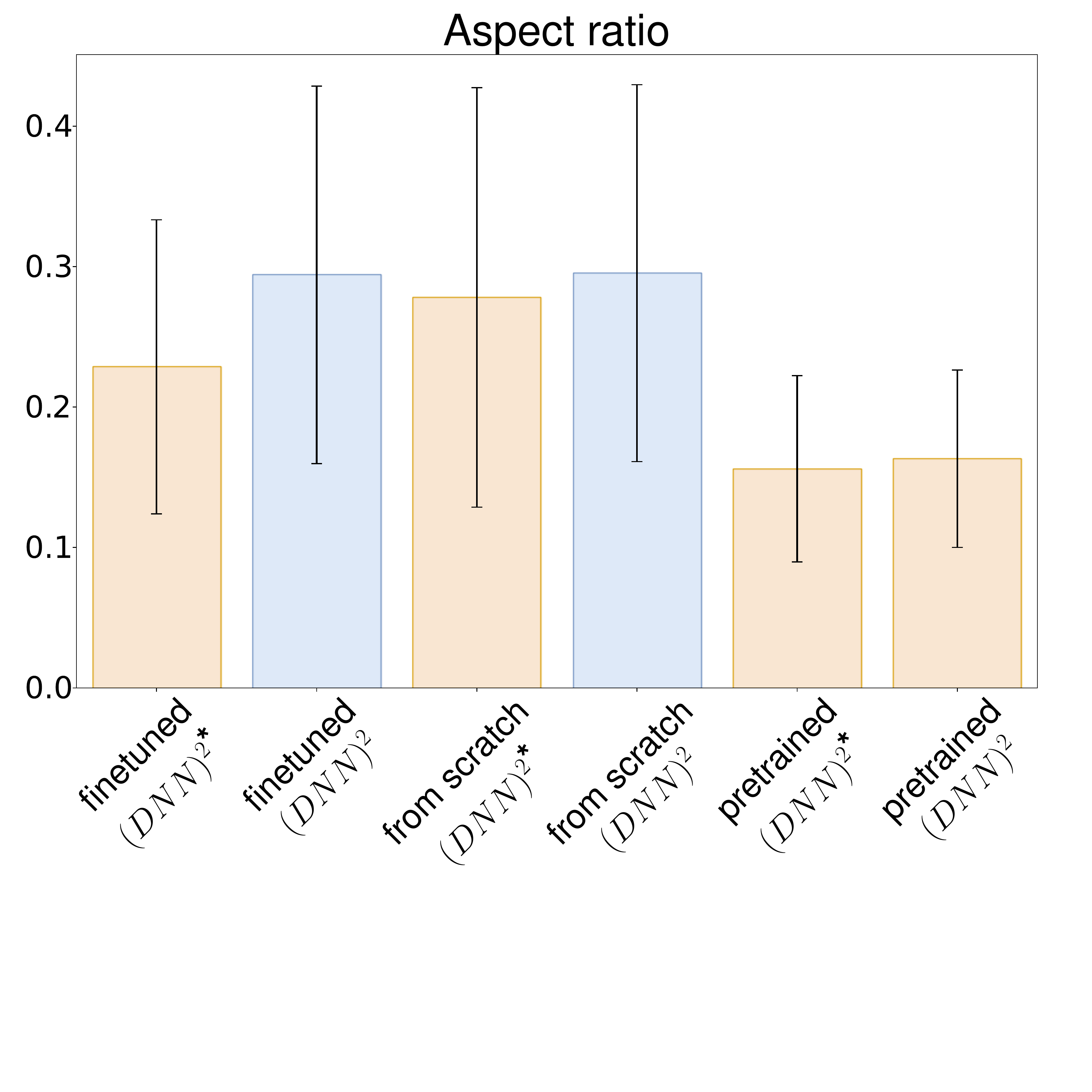} &
 \includegraphics[width=.31\columnwidth]{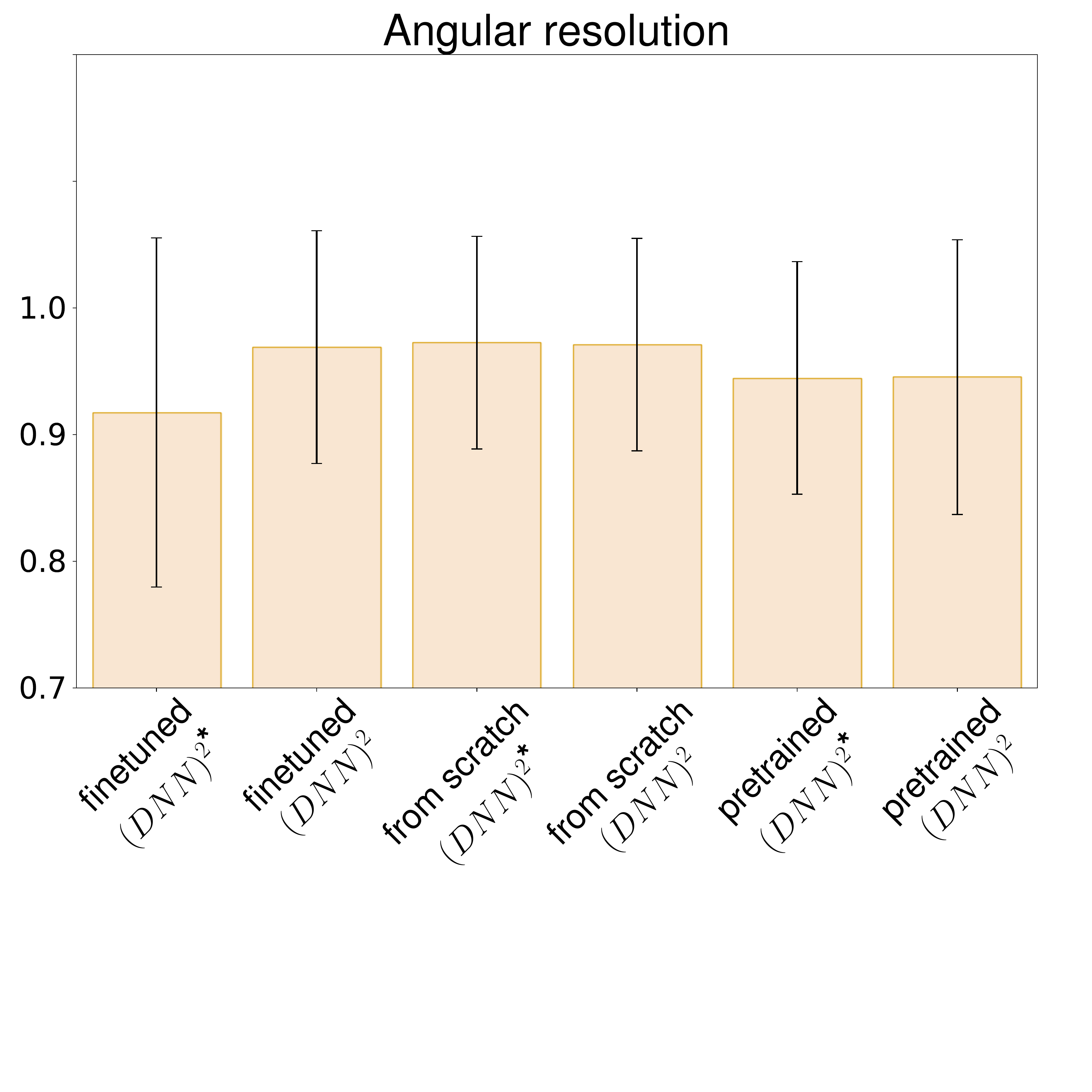}&
 \includegraphics[width=.31\columnwidth]{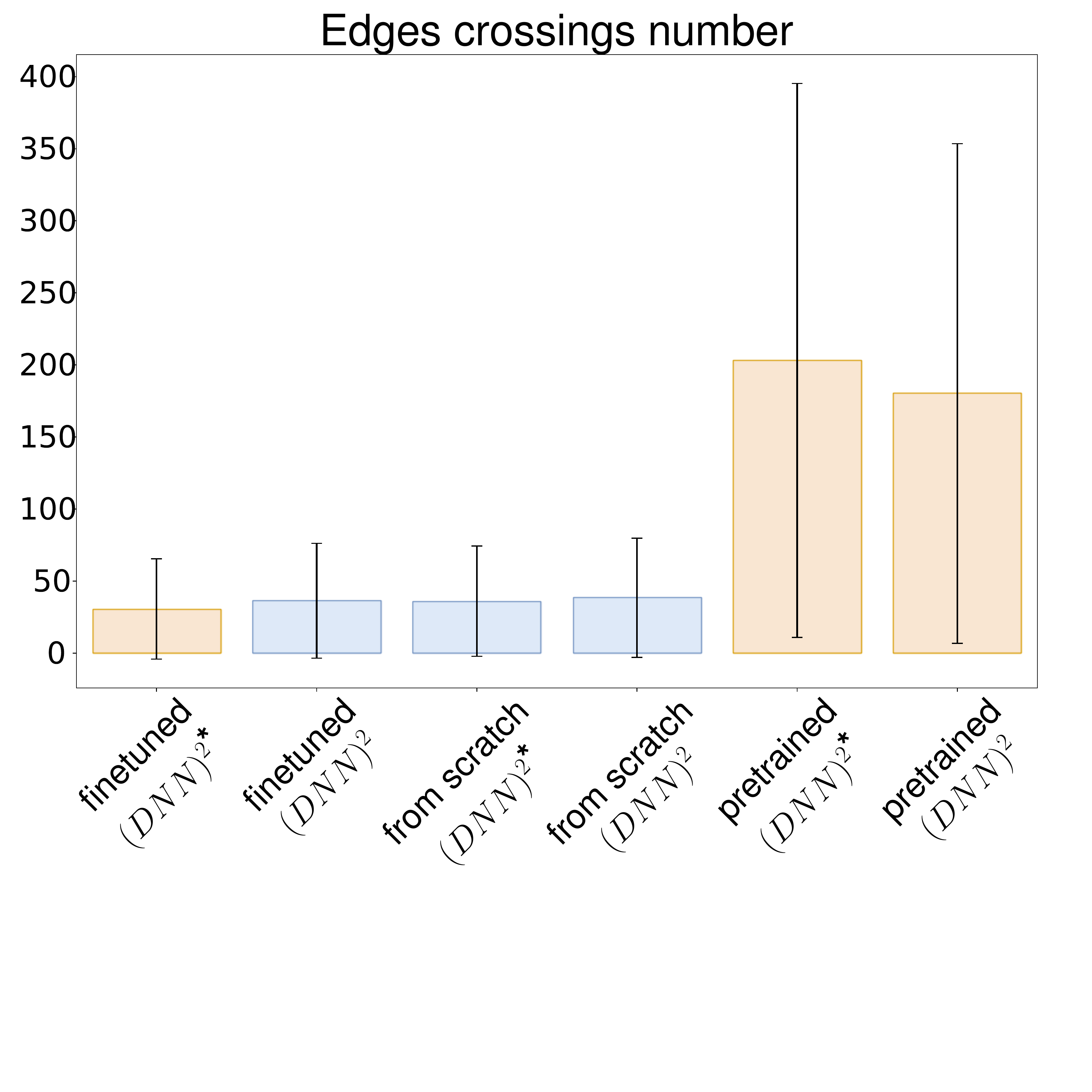}\\
 \includegraphics[width=.31\columnwidth]{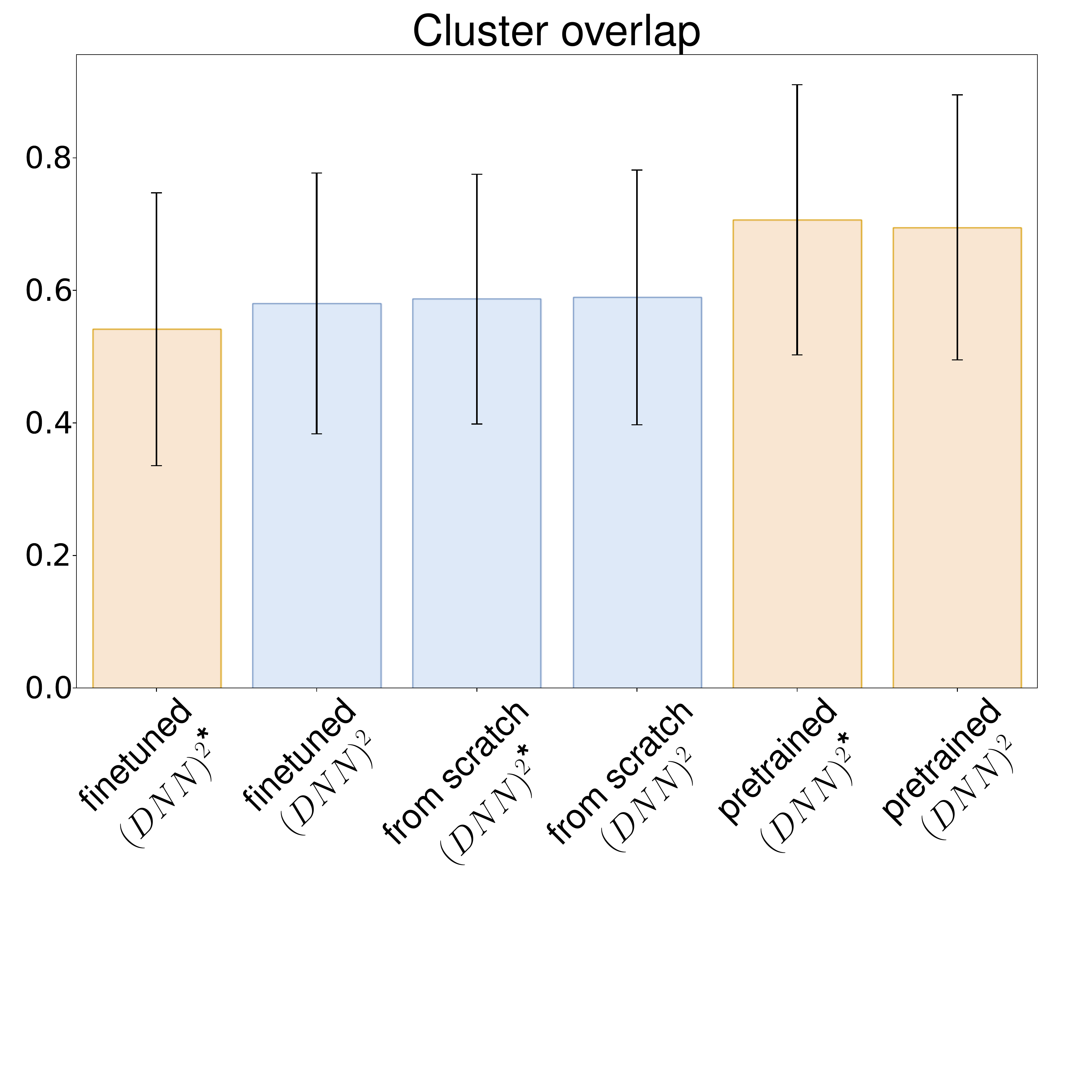}&
 \includegraphics[width=.31\columnwidth]{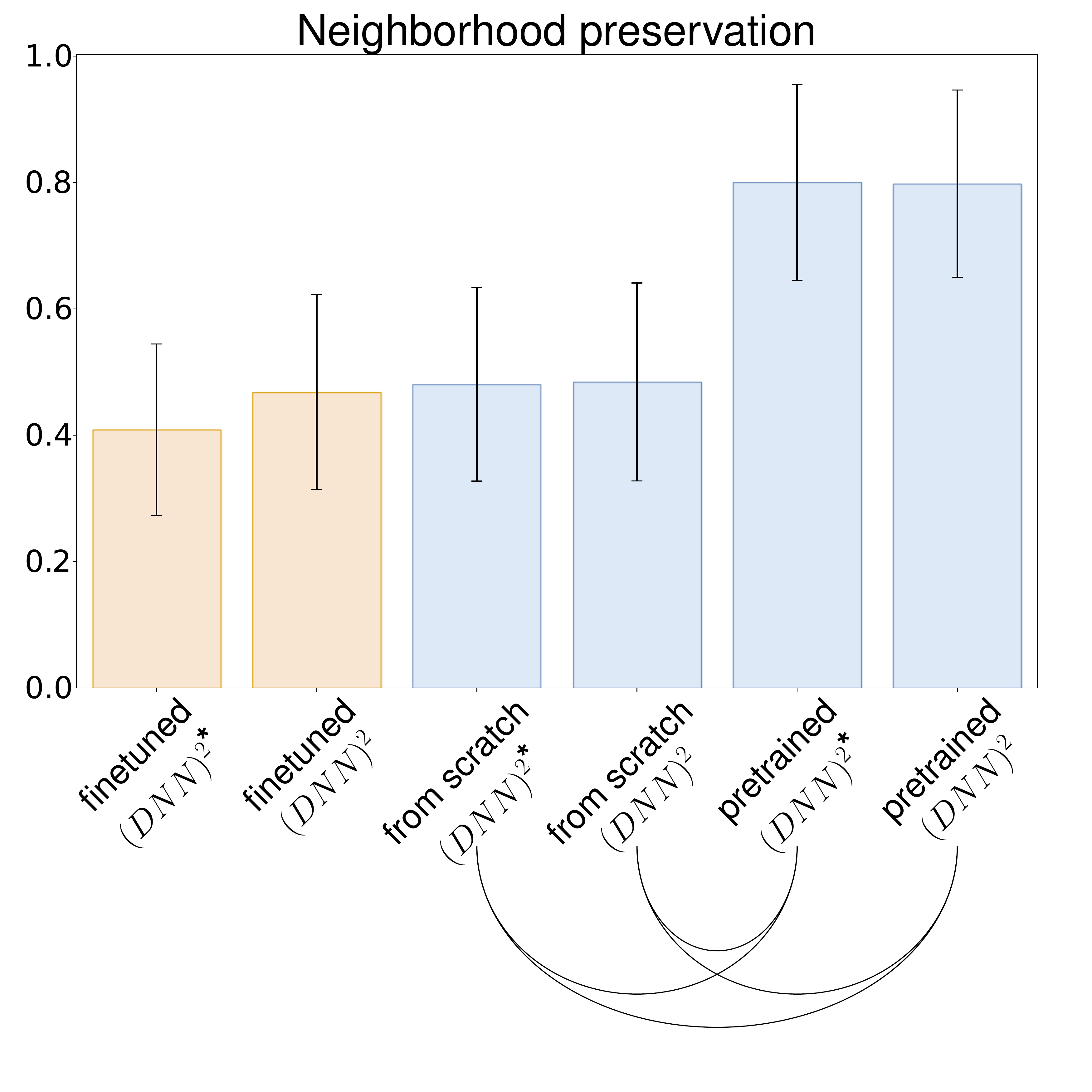}&
 \includegraphics[width=.31\columnwidth]{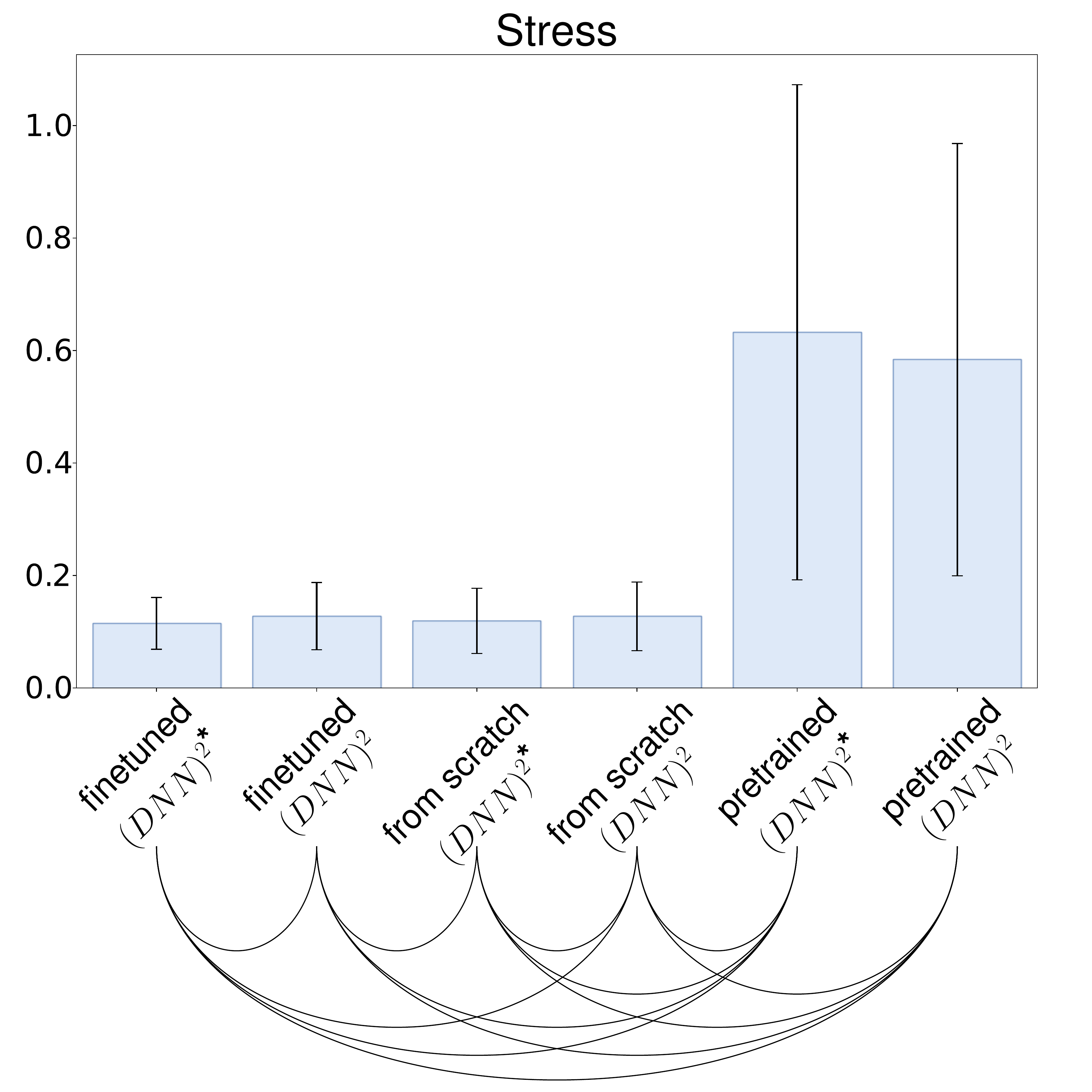}\\
\end{tabular}
}
\caption{Comparison of \modelname instances on the Rome test set. An orange bar indicates that the corresponding model performance is significantly different to \textit{all} others. An arc between two blue bars indicates pairwise significance.
}
\label{fig:benchmark1}
\end{figure}

This section compares the $6$ variants of \modelname to determine which training method is the most beneficial for the model. Execution times are not studied here and \autoref{fig:benchmark1} presents other metrics averages and standard deviations on the Rome test set for each \modelname instance. An orange bar indicates that the corresponding model performance is significantly different to \textit{all} others. An arc between two blue bars indicates that the difference of their performance is statistically significant.

\textit{Pretrained} instances perform significantly worse than others on all metrics but aspect ratio where they lead by a fair margin. \textit{From scratch} instances never perform the best on any metric. Overall, \textit{finetuned} \modelname lead to better scores with most metrics. It could be expected as it is well known in the Image Processing community that initialize weights to \textit{pretrained} values tends to speed up the training process and to lead to better performances including generalization to unseen data. The idea is that it is easier for the model to learn to solve a specific task if it already knows high-level features. 
As \textit{finetuned} models results are best, pretraining effectively learned the model such features that helped it to finetune.

In the next, only \textit{finetuned} instances of \modelname are compared to state-of-the-art methods since they perform better on the graph drawing task.

%%%%%%%%%%%%%%%%%%%%%%%%%%%%%%%%%%%%%%%%%%%%%%%%%%%%%%%%%%%%%%%%%%%%%%%%%%%%%%%
%%%%%%%%%%%%%%%%%%%%%%%%%%%%%%%%%%% BENCH 2 %%%%%%%%%%%%%%%%%%%%%%%%%%%%%%%%%%%
%%%%%%%%%%%%%%%%%%%%%%%%%%%%%%%%%%%%%%%%%%%%%%%%%%%%%%%%%%%%%%%%%%%%%%%%%%%%%%%
\subsection{Comparison with \textit{tsNET}}
\label{sec:benchmark2}

In this section, we study how the \modelname adaptation of $tsNET$ loss performs compared to the original Optimization-based implementation of $tsNET$ as both use the same cost function. The results are presented in \autoref{tab:benchmark2}.

\begin{table}[!t]
\centering
\resizebox{\columnwidth}{!}{
\begin{tabular}{|r|c|c|c|c|c|c|c|}
\cline{2-8}
\multicolumn{1}{c|}{} & \textbf{Aspect ratio} & \textbf{Angular res.} & \textbf{Cross. number} & \textbf{Cluster overlap} & \textbf{Neighb. preserv.} & \textbf{Stress} & \textbf{Exec. time} \\
\hline
\textbf{\modelname} &$0.3 \pm 0.136$ &$0.963 \pm 0.101$ &$34.9 \pm 40.5$ &$0.55 \pm 0.207$ &$0.451 \pm 0.159$ &\cellcolor{mgrey}$0.117 \pm 0.041$ &\cellcolor{mgrey}$20.4 \pm 10.4$ \\
\hline
$\boldsymbol{\mathit{tsNET}}$&\cellcolor{mgrey}$\mathbf{0.191 \pm 0.091}$ &\cellcolor{mgrey}$\mathbf{0.885 \pm 0.155}$ &\cellcolor{mgrey}$\mathbf{27.7 \pm 31.8}$ &\cellcolor{mgrey}$\mathbf{0.489 \pm 0.229}$ &\cellcolor{mgrey}$\mathbf{0.407 \pm 0.1}$ &$\mathbf{0.144 \pm 0.155}$ &$\mathbf{6541 \pm 5068}$ \\
\hline
\multicolumn{8}{c}{}\\ %emtpy space
\hline
\textbf{\modelnameStar} &$0.229 \pm 0.104$ &$0.905 \pm 0.15$ &$30.0 \pm 35.7$ &$0.507 \pm 0.214$ &$0.397 \pm 0.14$ &\cellcolor{mgrey}$0.111 \pm 0.042$ &\cellcolor{mgrey}$24.8 \pm 8.61$ \\
\hline
$\boldsymbol{\mathit{tsNET^*}}$ &$0.206 \pm 0.10$ &\cellcolor{mgrey}$\mathbf{0.872 \pm 0.181}$ &$27.1 \pm 32.0$ &\cellcolor{mgrey}$\mathbf{0.49 \pm 0.218}$ &\cellcolor{mgrey}$\mathbf{0.386 \pm 0.115}$ &$\mathbf{0.124 \pm 0.049}$ &$\mathbf{5836 \pm 5933}$ \\
\hline
\end{tabular}
}
\vspace{0.1cm}
\caption{\modelname and \modelnameStar pair comparison with their respective $tsNET$ algorithm on the Rome test set. Bold $tsNET$ scores are significantly different from their corresponding \modelname. The best \textit{significant} results are highlighted (\ie no highlight when the difference is \textit{not} significant).}
\label{tab:benchmark2}
\end{table}
\modelname has significantly lower scores than $tsNET$ on all metrics but Stress and Execution time. Though it is significant, the difference on Edge Crossings Number is small. The Execution time difference is heavily in favor of \modelname: $20.4ms$ as opposed to $6541ms$ for $tsNET$. The trends are about the same between \modelnameStar and $tsNET^*$, but their scores are closer and the differences in Angular Resolution and Edge Crossings Number are not significantly different anymore. \modelnameStar is also better on Stress and strongly better on Execution time.

It is noteworthy that $tsNET$ and $tsNET^*$ suffers from a significantly high Execution time standard deviation, meaning that the methods hardly converge on some graphs. In addition, \num{450} out of \num{1931} (\ie 23\%) graphs were excluded from the test set as the $tsNET^*$ implementation would not complete on these.

We can conclude that the \modelname implementation that adapts $tsNET$ to a Deep Learning approach is faster but does not lead to better drawings according to most of the metrics. Although quality metrics differences are significant, they remain small and should undeniably be alleviated by future works.

$tsNET$ is designed to optimize a specific input graph at a time whereas, with the Deep Learning approach, we aim at optimizing the model and not the drawing of a single graph. If the DL training process is computationally expensive, the resulting model should be capable of computing the layout without any further need for optimization. In fact, if a DL model learns well to lay graphs out by optimizing a generic cost function, it suggests that there exists a bounded sequence of operations that efficiently projects a graph in a 2D space.

\begin{table}[!t]
\centering
\resizebox{\columnwidth}{!}{
\begin{tabular}{|r|r|r|r|r|r|r|r|}
\cline{2-8}
\multicolumn{1}{c|}{} & \multicolumn{1}{c|}{\textbf{Aspect ratio}} & \multicolumn{1}{c|}{\textbf{Angular res.}} & \multicolumn{1}{c|}{\textbf{Cross. number}} & \multicolumn{1}{c|}{\textbf{Cluster overlap}} & \multicolumn{1}{c|}{\textbf{Neighb. preserv.}} & \multicolumn{1}{c|}{\textbf{Stress}} & \multicolumn{1}{c|}{\textbf{Exec. time}} \\
\hline
\textbf{\modelname \textsuperscript{o}} &$0.294 \pm 0.134$ \textsuperscript{+} &$0.969 \pm 0.092$ \textsuperscript{+} &$36.3 \pm 39.9$ \textsuperscript{+} &$0.58 \pm 0.197$ \textsuperscript{+} &$0.468 \pm 0.154$ \textsuperscript{+} &$0.128 \pm 0.06$ \textsuperscript{+} & $21.0 \pm 10.3$ \textsuperscript{+} \\
\hline
\textbf{\modelnameStar \textsuperscript{+}} &\cellcolor{mgrey}$0.229 \pm 0.105$ \textsuperscript{o}&$0.917 \pm 0.138$ \textsuperscript{o}&\cellcolor{mgrey}$30.6 \pm 34.8$ \textsuperscript{o} &\cellcolor{mgrey}$0.541 \pm 0.206$ \textsuperscript{o} &\cellcolor{mgrey}$0.409 \pm 0.136$ \textsuperscript{o} &$0.115 \pm 0.046$ \textsuperscript{o} &$25.1 \pm 8.36$ \textsuperscript{o} \\
\hline
\hline
\textbf{t-SNE} &$0.276 \pm 0.158$ \textsuperscript{o+} &$0.97 \pm 0.038$ \textsuperscript{o+} &$69.1 \pm 48.9$ \textsuperscript{o+} &$0.598 \pm 0.252$ \textsuperscript{o+} &$0.584 \pm 0.097$ \textsuperscript{o+} &$0.56 \pm 0.771$ \textsuperscript{o+} &$166 \pm 71.7$ \textsuperscript{o+} \\
\hline
\textbf{PivotMDS} &$0.298 \pm 0.125$ \textsuperscript{+} &$0.978 \pm 0.088$ \textsuperscript{o+} &$38.7 \pm 43.6$ \textsuperscript{+} &$0.623 \pm 0.202$ \textsuperscript{o+} &$0.49 \pm 0.17$ \textsuperscript{o+}&$0.104 \pm 0.035$ \textsuperscript{o+} &\cellcolor{mgrey}$0.546 \pm 0.478$ \textsuperscript{o+} \\
\hline
\textbf{GEM} &$0.573 \pm 0.197$ \textsuperscript{o+} &$0.972 \pm 0.034$ \textsuperscript{o+} &$54.4 \pm 61.2$ \textsuperscript{o+} &$0.722 \pm 0.162$ \textsuperscript{o+} &$0.617 \pm 0.123$ \textsuperscript{o+} &$0.24 \pm 0.062$ \textsuperscript{o+} &$5.22 \pm 3.83$ \textsuperscript{o+} \\
\hline
\textbf{\sgdgd} &$0.263 \pm 0.123$ \textsuperscript{o+} &\cellcolor{mgrey}$0.812 \pm 0.208$ \textsuperscript{o+} &\cellcolor{mgrey}$32.2 \pm 36.6$ \textsuperscript{o} &$0.583 \pm 0.204$ \textsuperscript{+} &$0.439 \pm 0.181$ \textsuperscript{o+} &\cellcolor{mgrey}$0.066 \pm 0.027$ \textsuperscript{o+} &$1.13 \pm 0.91$ \textsuperscript{o+} \\
\hline
\end{tabular}
}
\vspace{0.1cm}
\caption{\modelname and \modelnameStar pair comparisons with selected state-of-the-art algorithms. \textsuperscript{o} (resp. \textsuperscript{+}) indicates a significant difference with \modelname (resp. \modelnameStar). The best significant results are highlighted.}
\label{tab:benchmark3}
\end{table}
%%%%%%%%%%%%%%%%%%%%%%%%%%%%%%%%%%%%%%%%%%%%%%%%%%%%%%%%%%%%%%%%%%%%%%%%%%%%%%%
%%%%%%%%%%%%%%%%%%%%%%%%%%%%%%%%%%% BENCH 3 %%%%%%%%%%%%%%%%%%%%%%%%%%%%%%%%%%%
%%%%%%%%%%%%%%%%%%%%%%%%%%%%%%%%%%%%%%%%%%%%%%%%%%%%%%%%%%%%%%%%%%%%%%%%%%%%%%%

\subsection{Comparison with State-of-the-art Layout Algorithms}
\label{sec:benchmark3}

This section studies how \modelname performs compared to selected layout algorithms from the literature: t-SNE~\cite{tsne}, since we leverage the Kullback-Leibler divergence, PivotMDS~\cite{pivotmds}, a deterministic Multidimensional Scaling used by \modelnameStar and $tsNET^*$, GEM~\cite{frick1994gem}, a well-established force-directed technique
and \sgdgd~\cite{zheng2018sgdgd}, a \textit{stress} Optimization-based approach with SGD. The methods are compared on the Rome test set and the results are reported in \autoref{tab:benchmark3}. \modelname scores are slightly different from \autoref{tab:benchmark2} since all test graphs are taken into account here.

\modelnameStar performs better than \modelname as all aesthetic metrics are significantly in its favor. \modelnameStar is slower due to the extra processing of PivotMDS it requires. This outcome was expected in view of $tsNET$ variants comparisons in \cite{kruiger2017tsnet}.

\modelname is better than GEM on all quality metrics; and is significantly better than t-SNE on all metrics but Aspect ratio. It performs better than PivotMDS on Angular resolution, Cluster overlap and Neighborhood preservation, but is outperformed on Aspect ratio and Stress, while the difference is not significant on Edge Crossings Number. Finally, \sgdgd performs significantly better than \modelname on all metrics but Cluster overlap.

Overall, \modelnameStar is significantly better on Aspect ratio, Cluster overlap and Neighborhood preservation than all the other considered methods. It is also the best in Edge Crossings Number with \sgdgd. While it was observed to be better than $tsNET^{*}$ on Stress, it is here outperformed by PivotMDS and \sgdgd.

As for Execution time, we can see that both \modelname variants are slower than other methods except t-SNE. However, they are less sensible to graph size variations: \modelname variants execution time standard deviations are 33\% and 47\% of their average, while they range between 43\% and 87\% for other methods. It is important to note that a forward pass time in \modelname is almost constant and only takes $1.4ms$ (\ie 6\% of its total execution time), the remaining time being used to pre-process data for the model inputs.

Although \modelname is not the best performing variant, its results indicate that a Deep Learning framework, without any knowledge of what is a graph layout, can learn a sequence of operations that lays graphs out. \modelnameStar leveraged its PivotMDS input and drawn better layouts according to the quality metrics. Its performances make it a good trade-off between $tsNET^*$ and \sgdgd. The latter performed surprisingly well, while GEM underperformed in this evaluation.

\section{Discussion}
\label{sec:discussion}
\subsection{Visual evaluation}
\label{sec:visual_eval}

\begin{figure}[!tb]
\centering
\resizebox{\columnwidth}{!}{
\begin{tabular}{|c|c|c|c|c|}
\cline{2-5}
\multicolumn{1}{c|}{} & Dodecahedron & Grid & \makecell{Rome graph 1\\\textit{id}: 138} & \makecell{Rome graph 2\\\textit{id}: \num{10082}} \\
\hline
 \begin{sideways}{\hspace{0.85cm}\modelname}\end{sideways}& 
 {\includegraphics[width=.24\columnwidth]{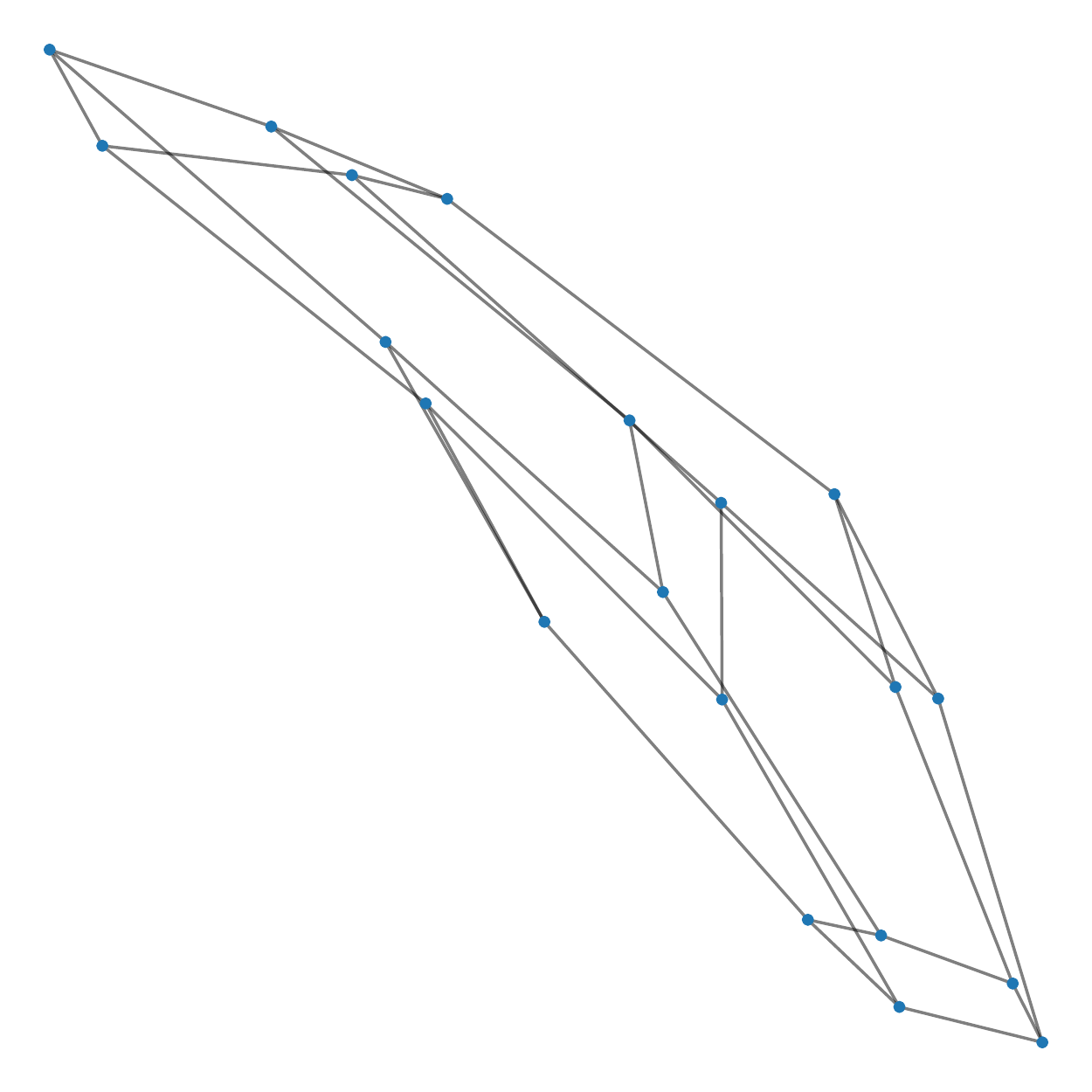}}&
 {\includegraphics[width=.24\columnwidth]{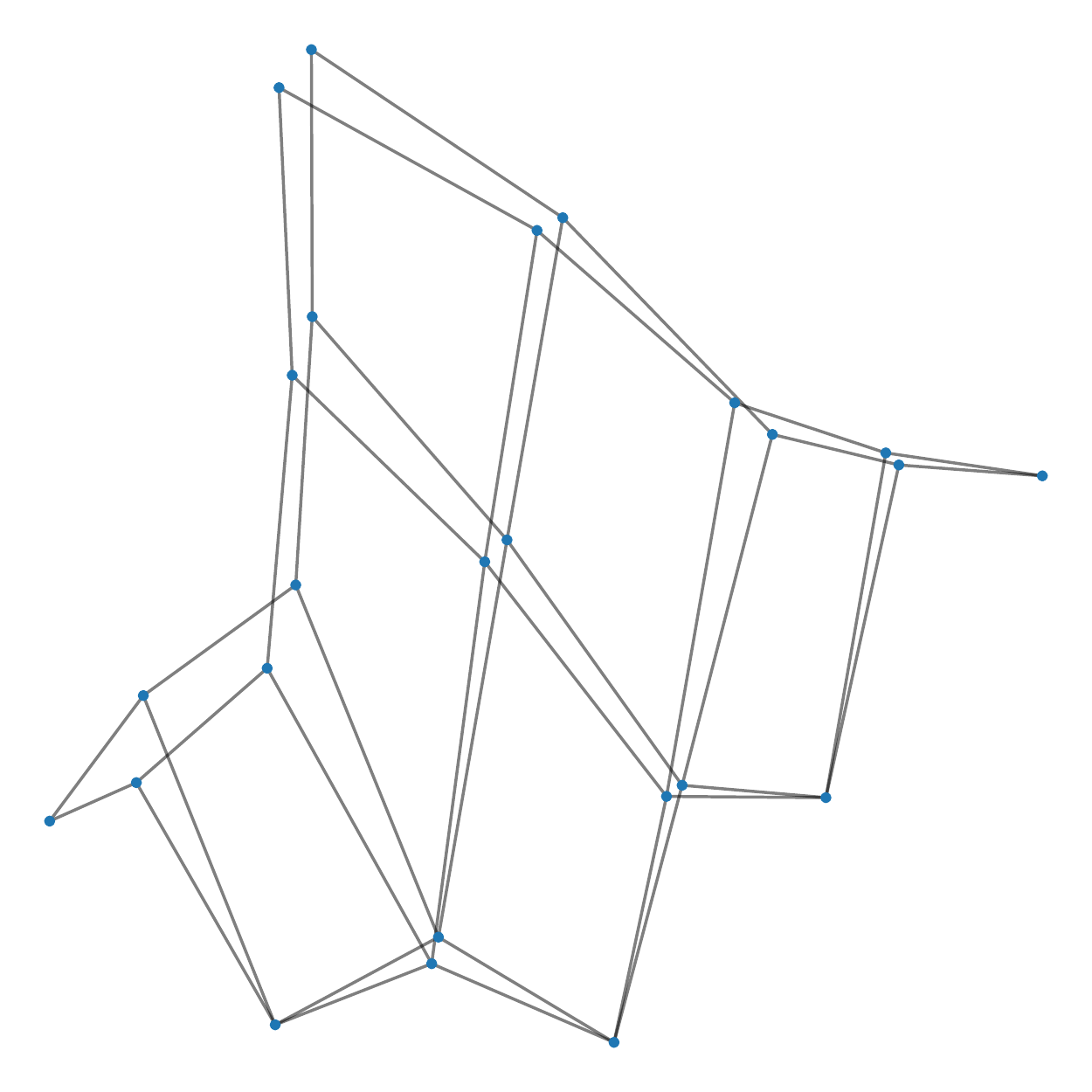}}&
 {\includegraphics[width=.24\columnwidth]{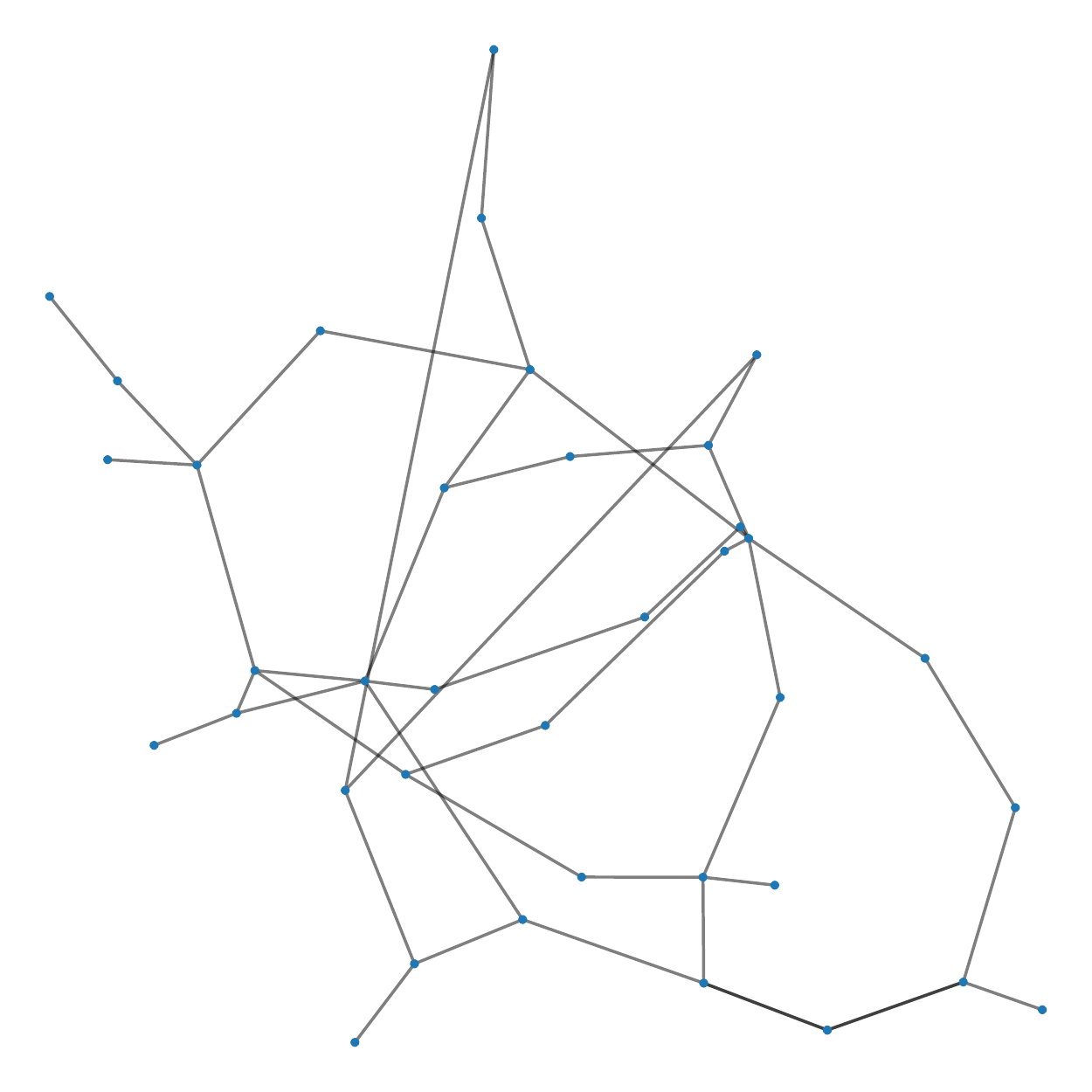}}&
 {\includegraphics[width=.24\columnwidth]{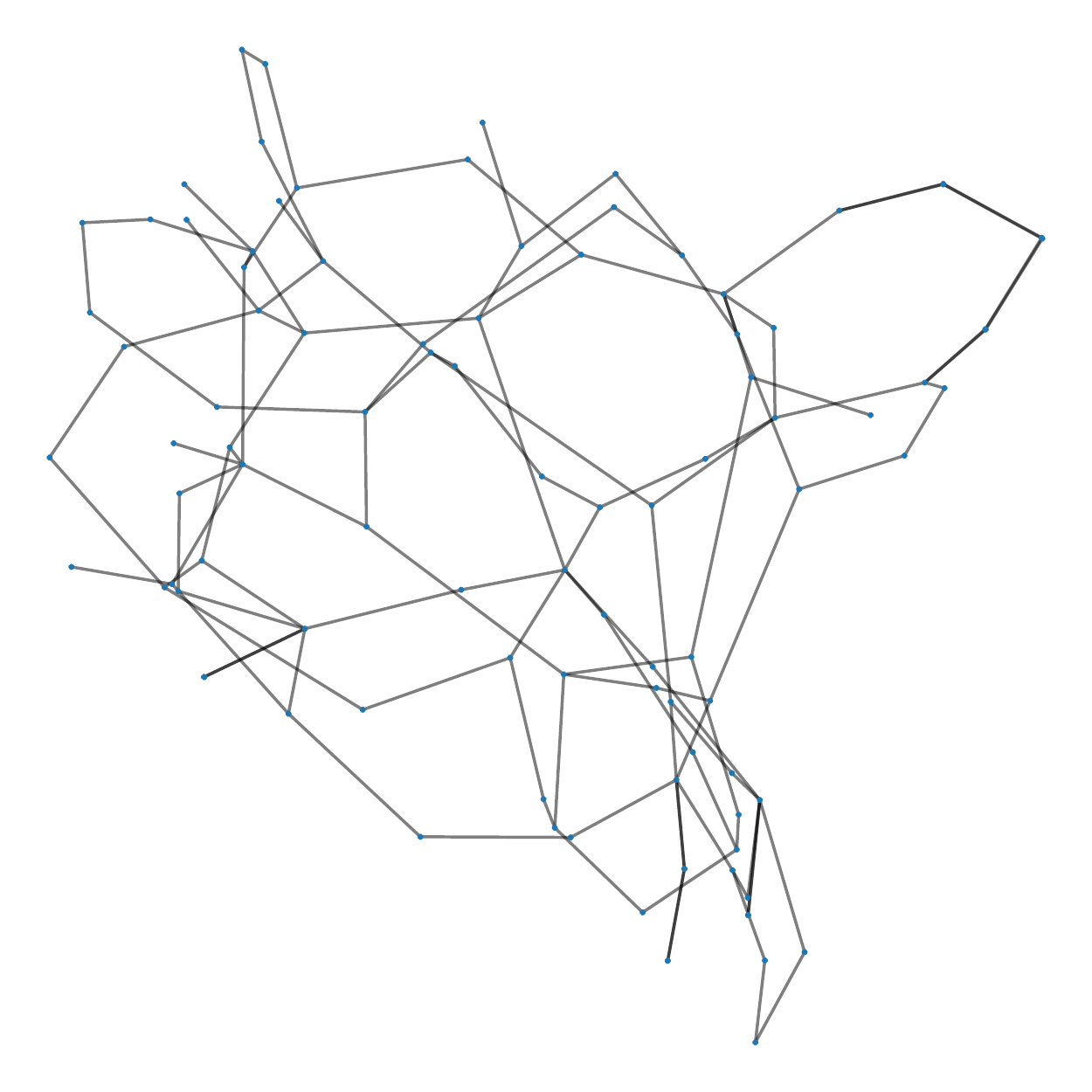}}\\
\hline
 \begin{sideways}{\hspace{0.85cm}\modelnameStar}\end{sideways}& 
 {\includegraphics[width=.24\columnwidth]{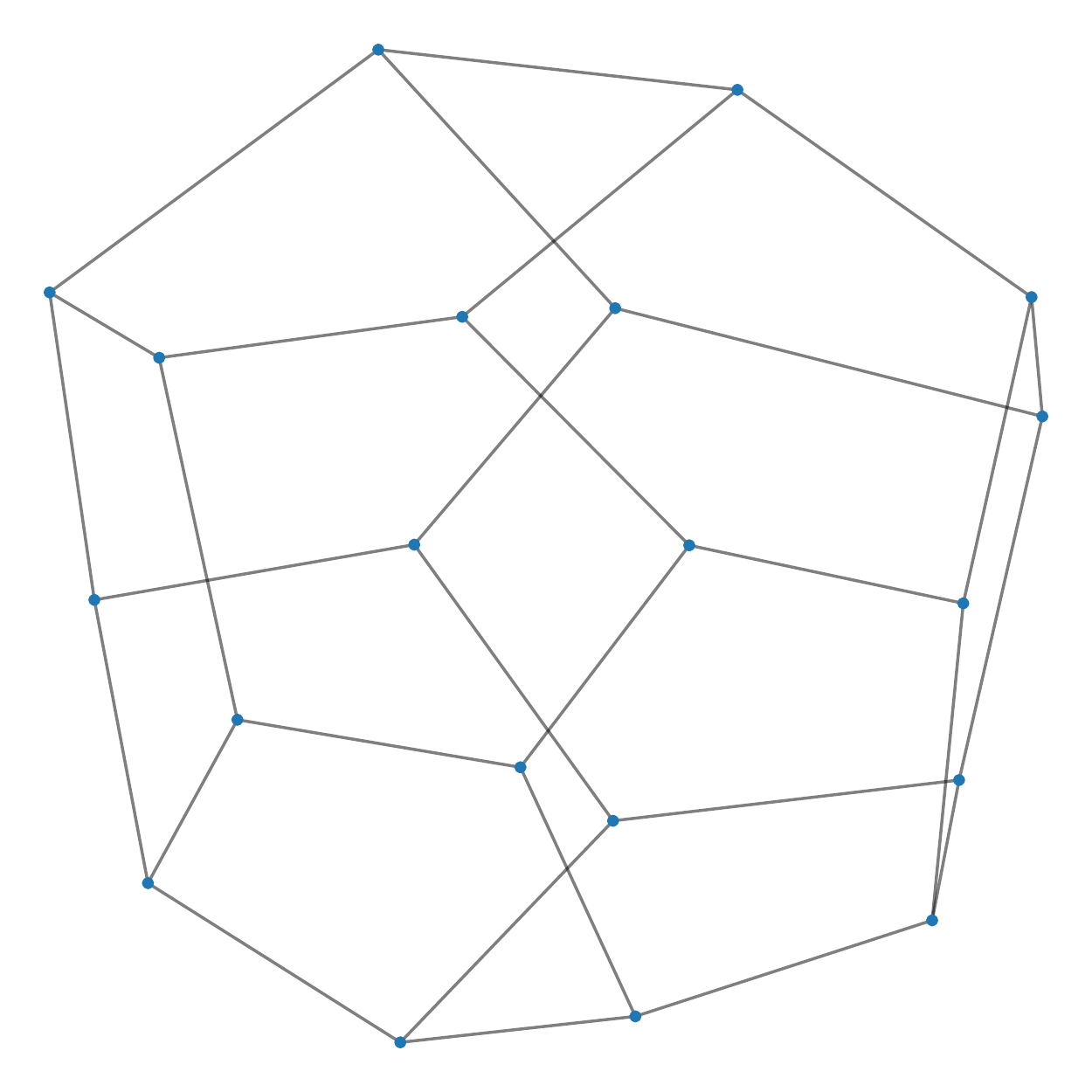}}&
 {\includegraphics[width=.24\columnwidth]{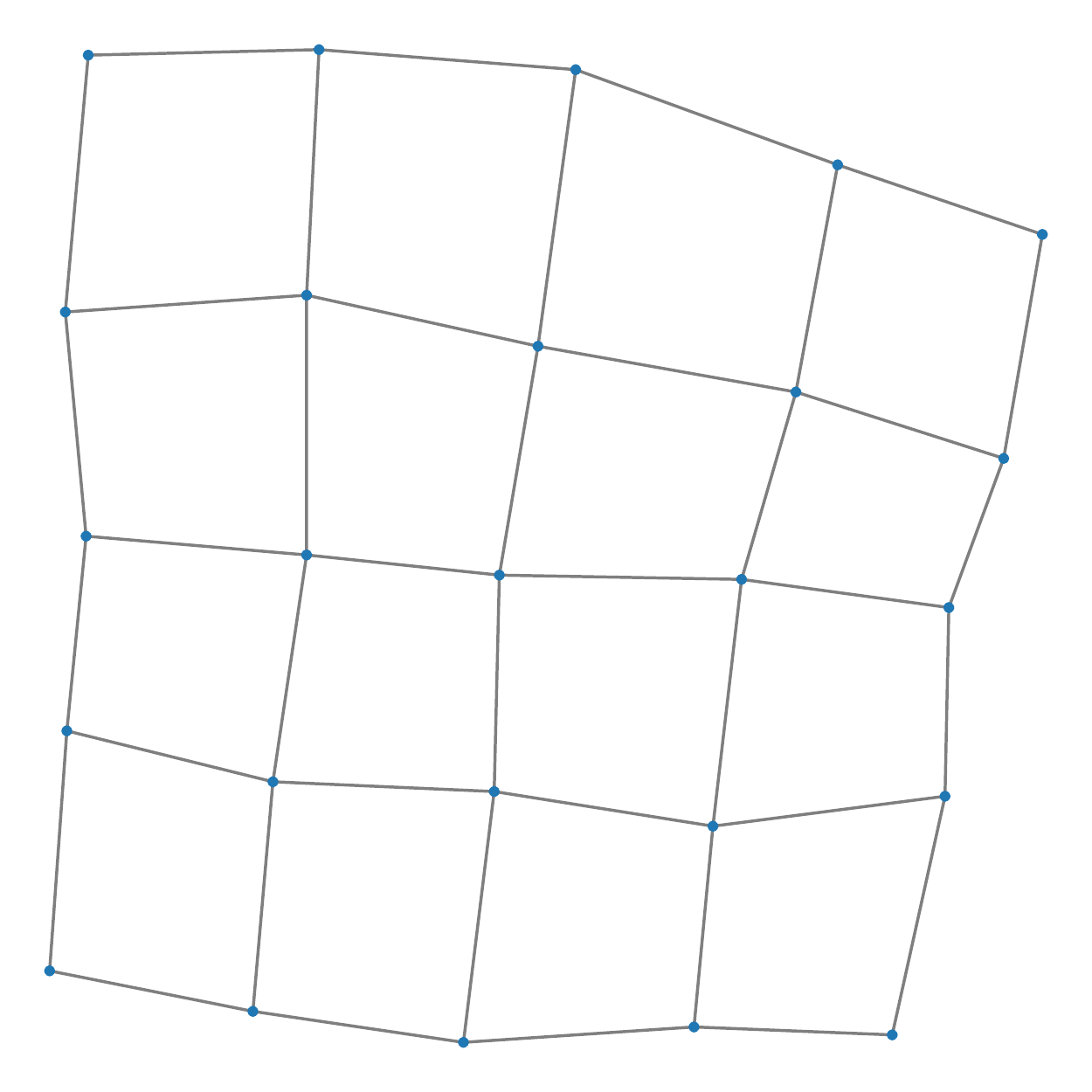}}&
 {\includegraphics[width=.24\columnwidth]{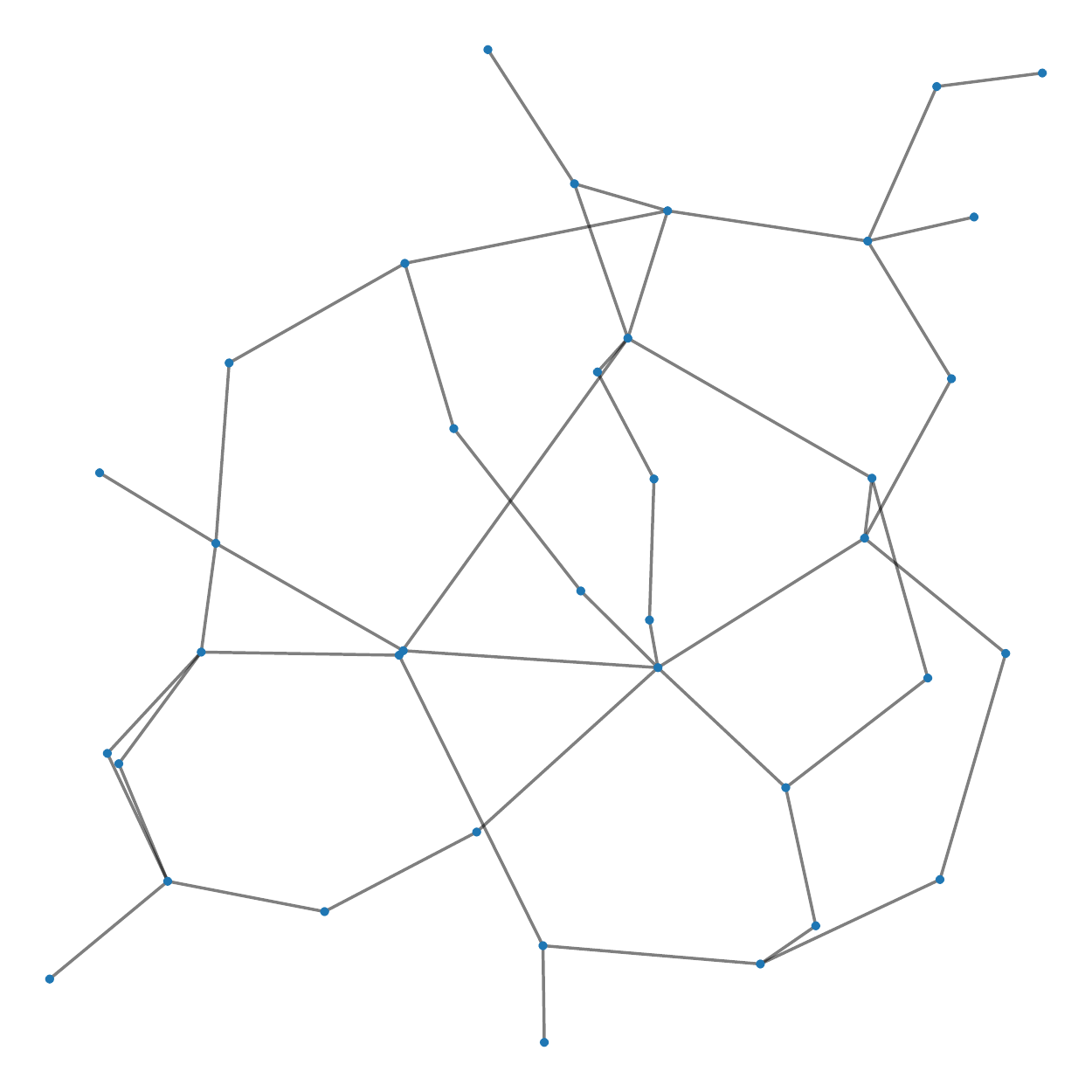}} &
 {\includegraphics[width=.24\columnwidth]{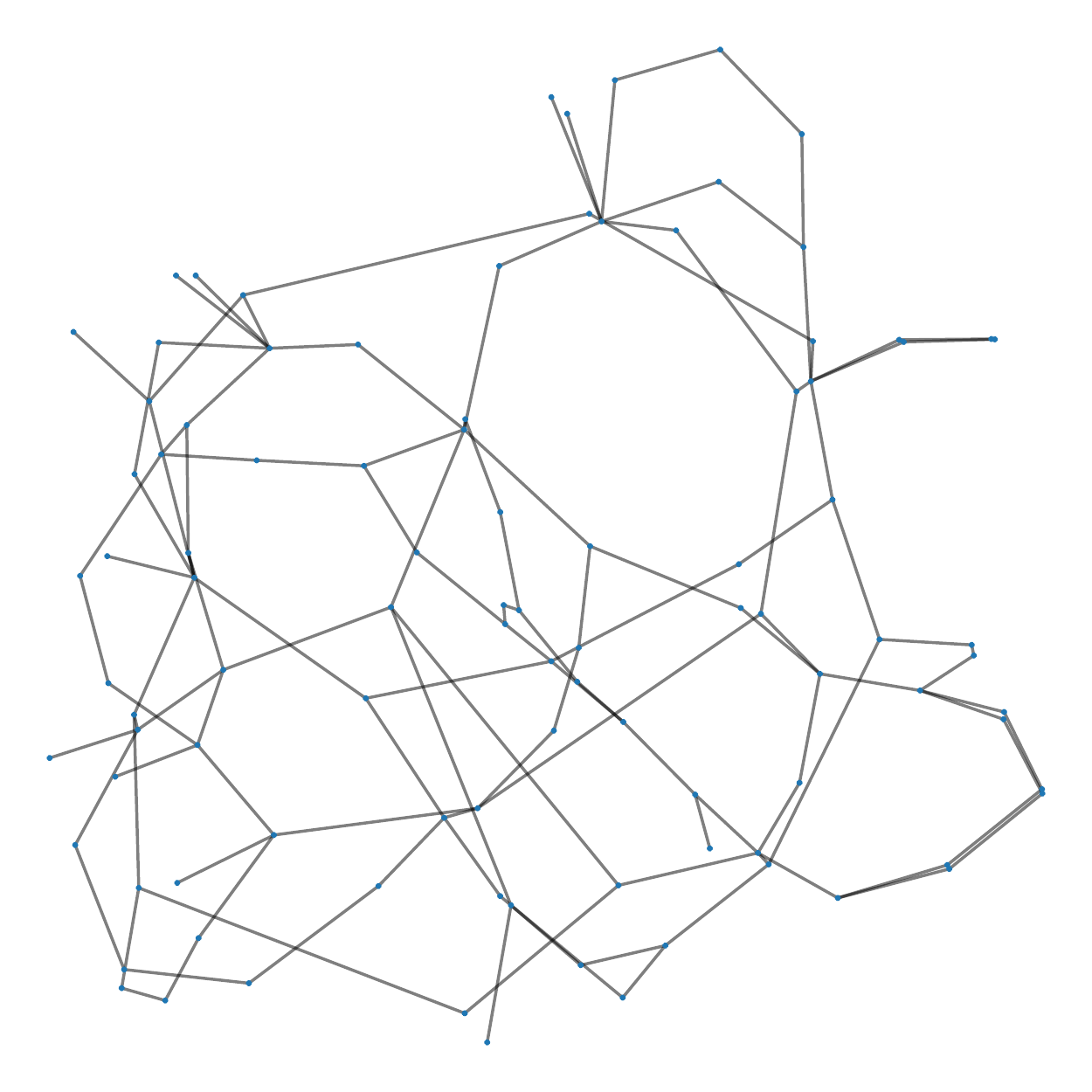}}\\
\hline
 \begin{sideways}{\hspace{0.9cm}$tsNET^{*}$}\end{sideways}& 
 {\includegraphics[width=.24\columnwidth]{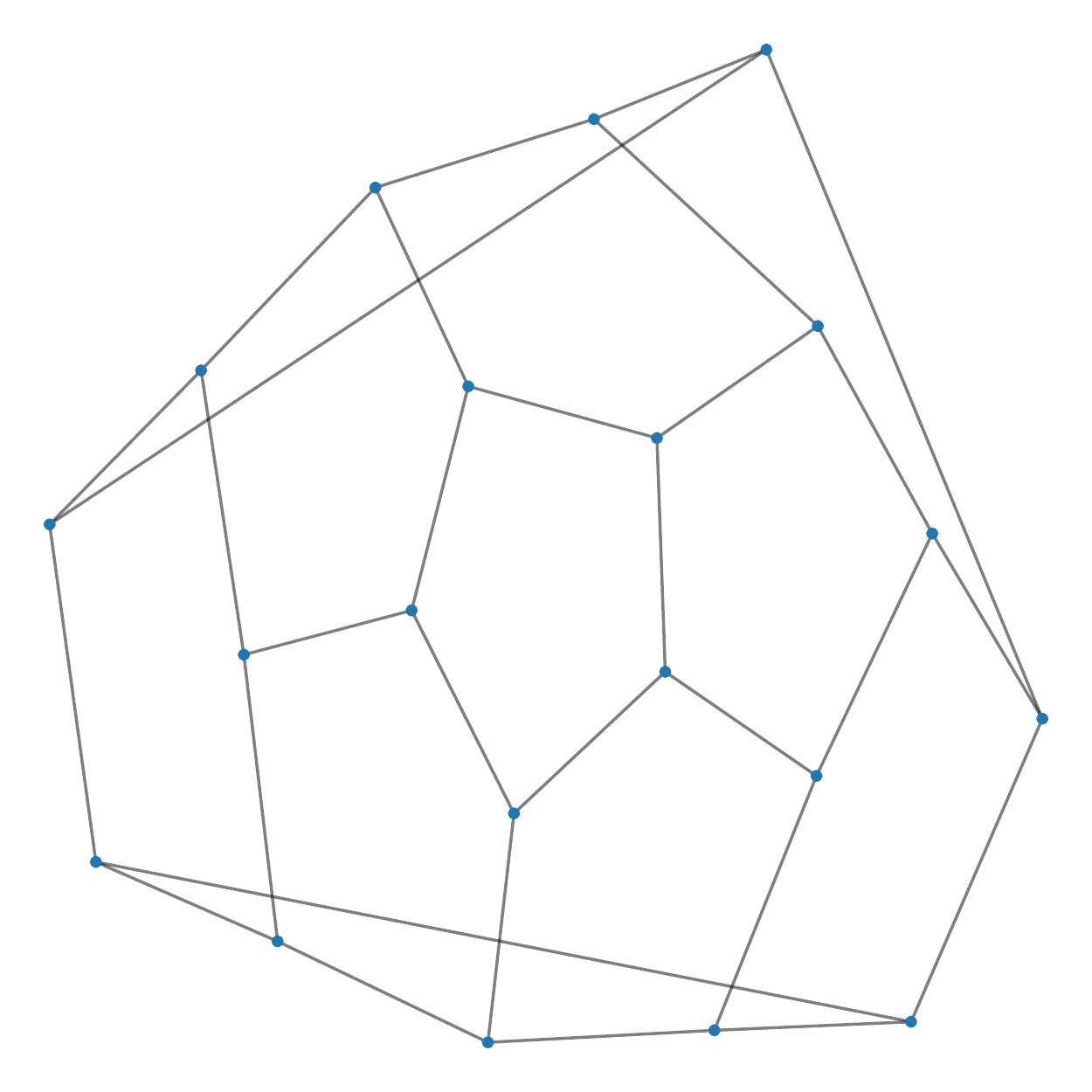}}&
 {\includegraphics[width=.24\columnwidth]{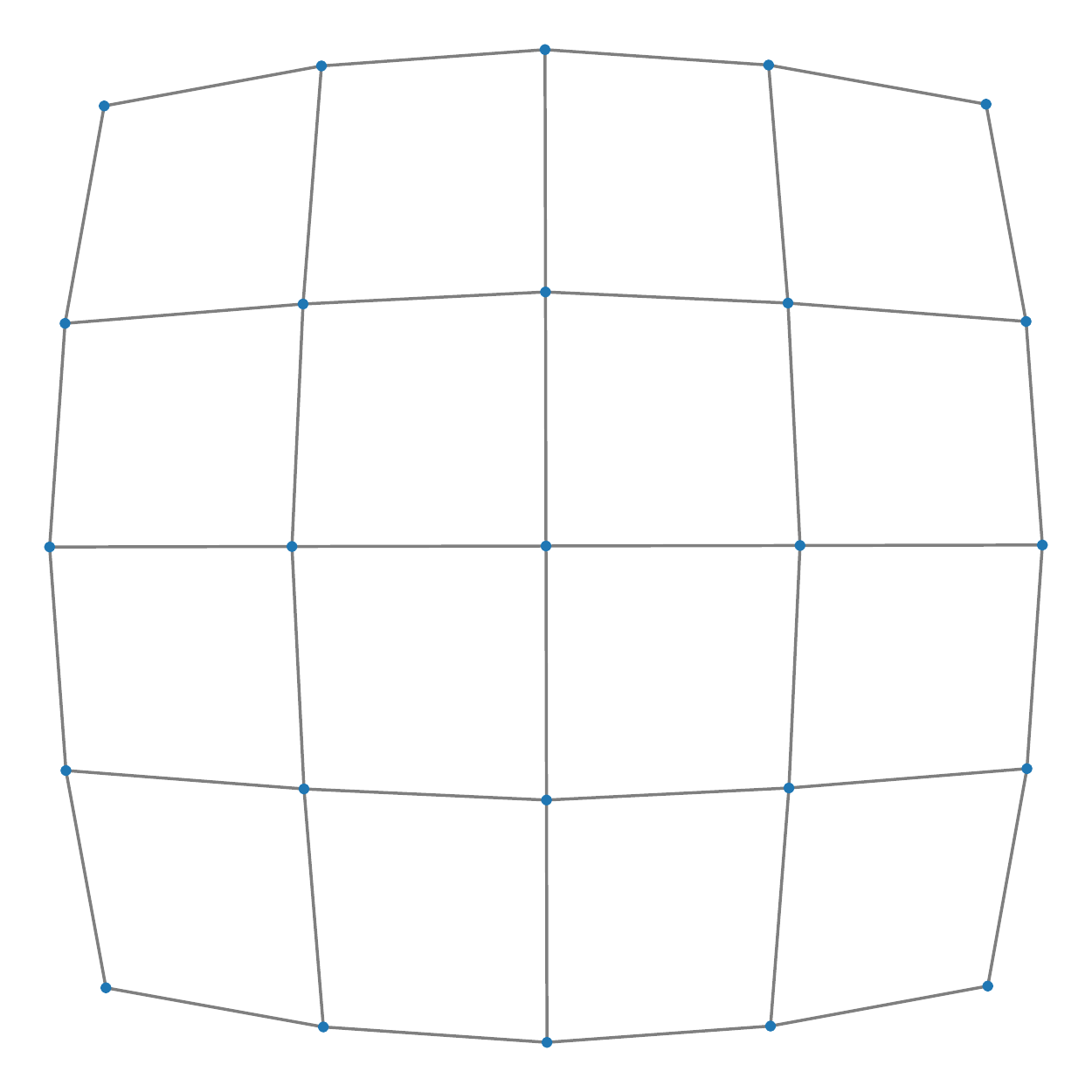}}&
 {\includegraphics[width=.24\columnwidth]{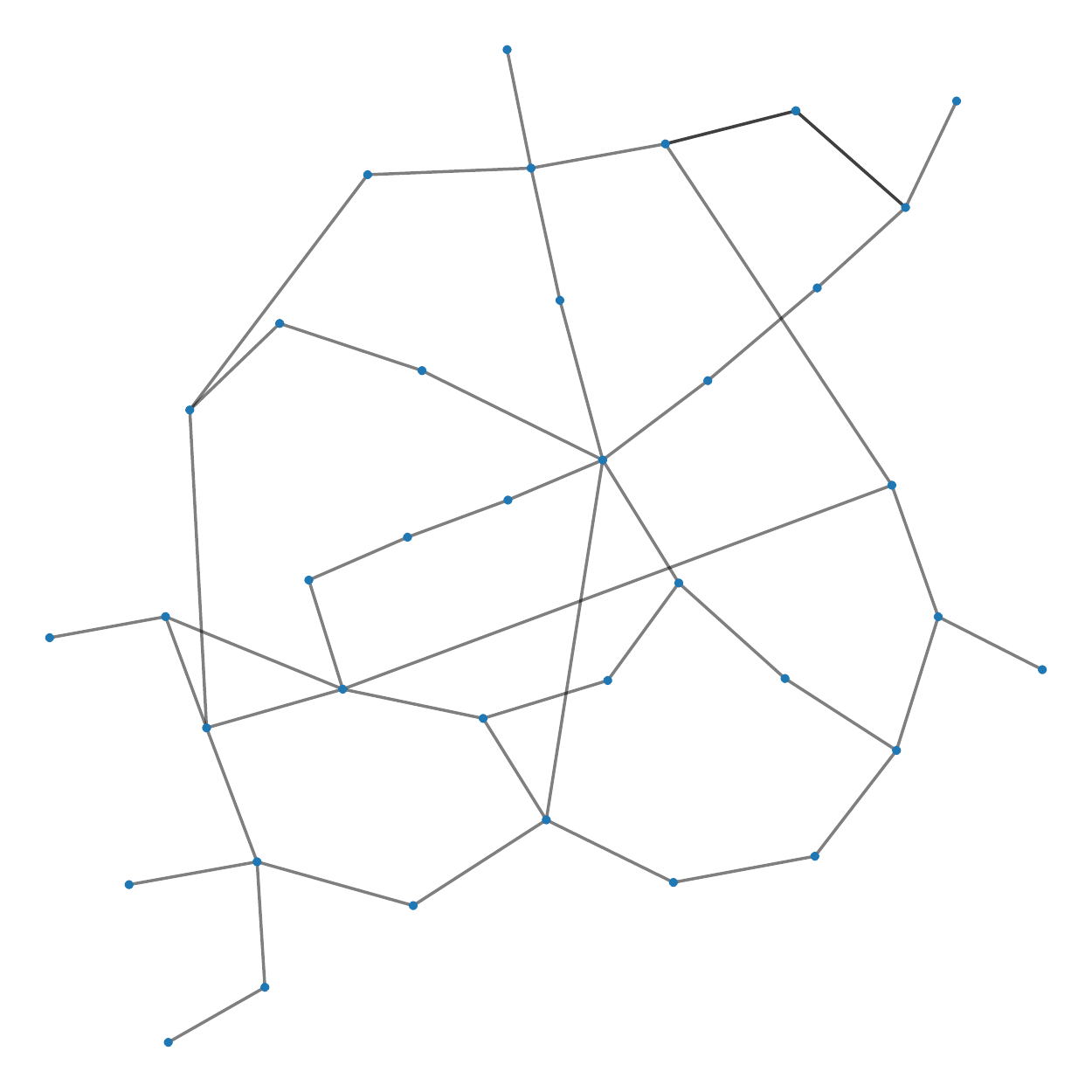}}&
 {\includegraphics[width=.24\columnwidth]{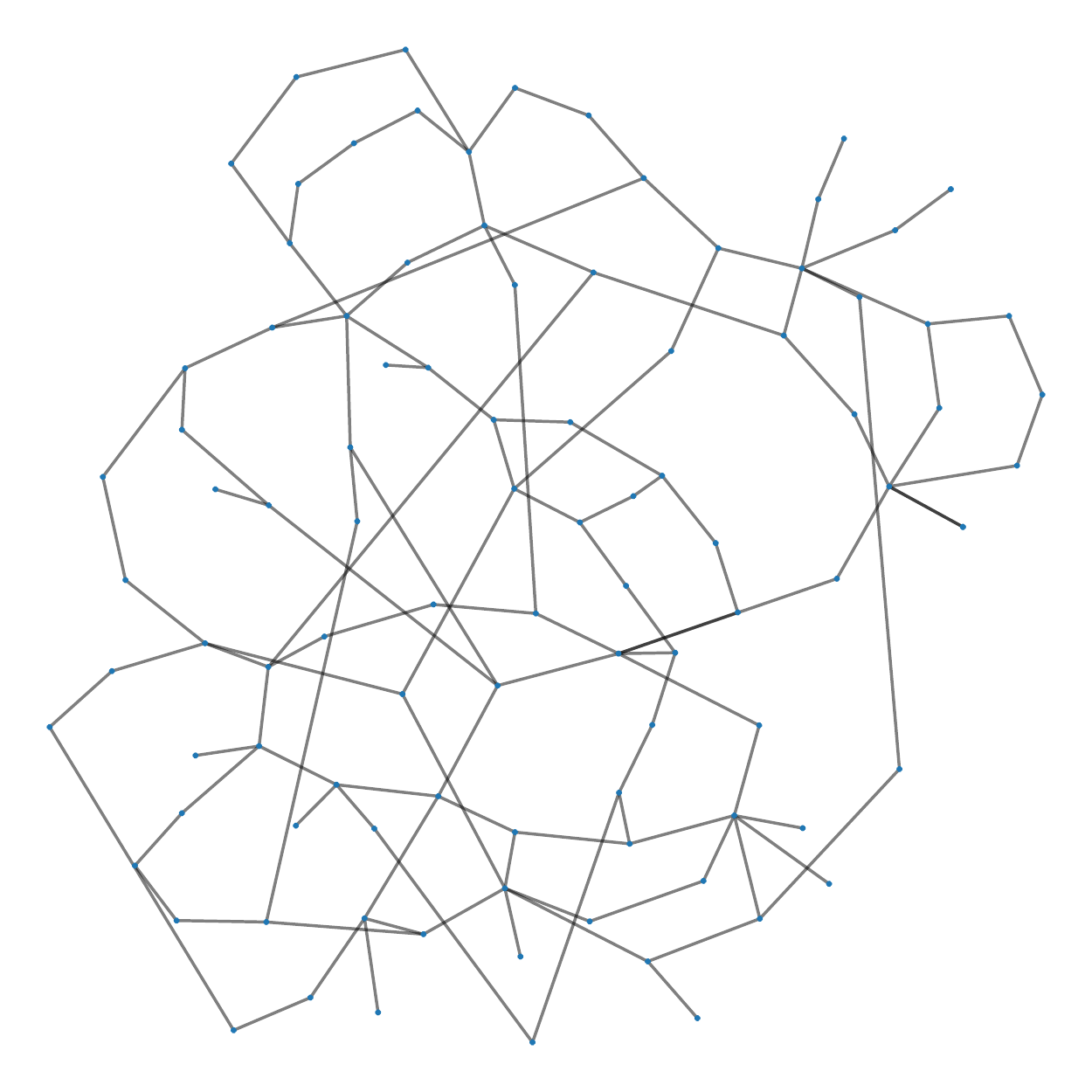}}\\
\hline
 \begin{sideways}{\hspace{1.1cm}\sgdgd}\end{sideways}& 
  {\includegraphics[width=.24\columnwidth]{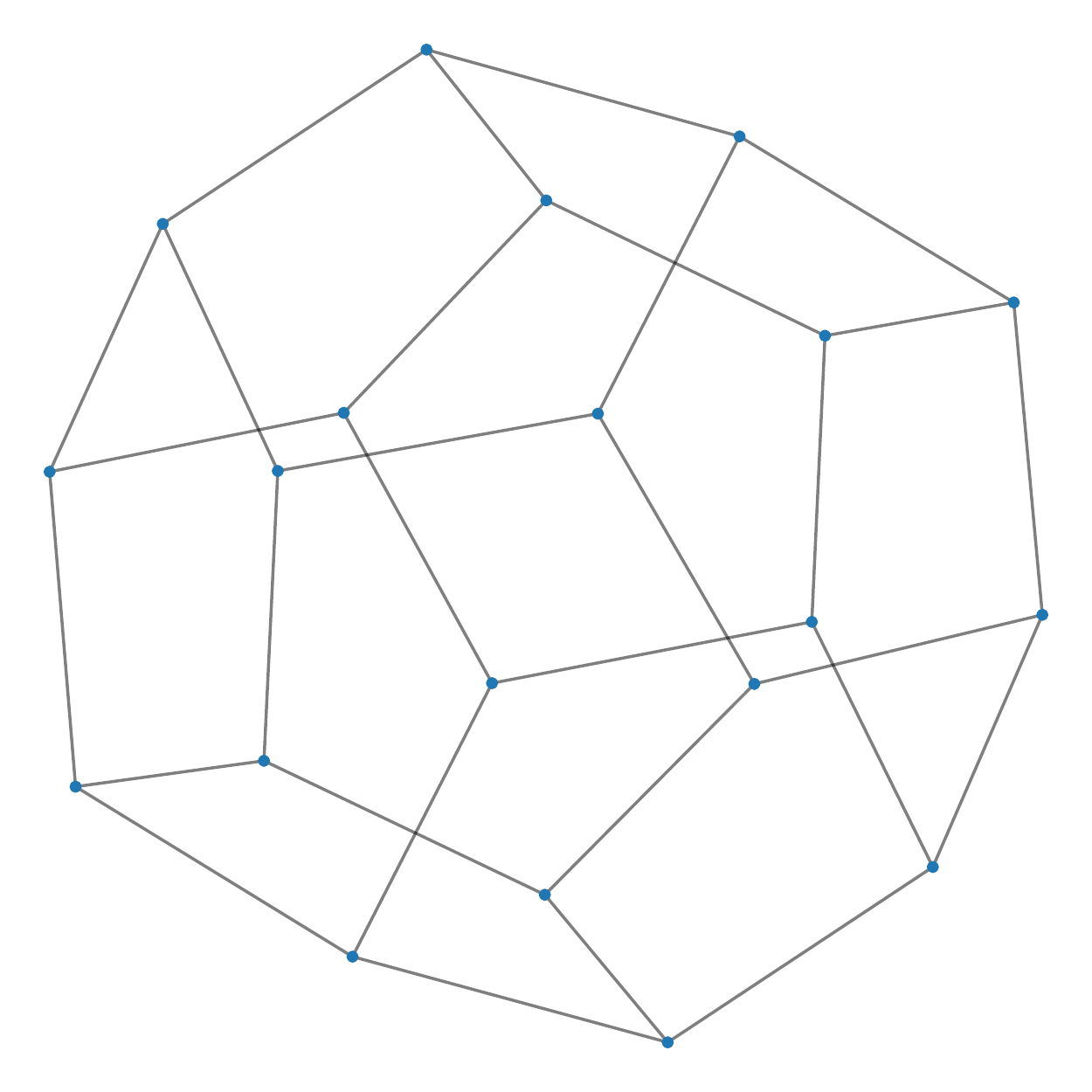}}&
  {\includegraphics[width=.24\columnwidth]{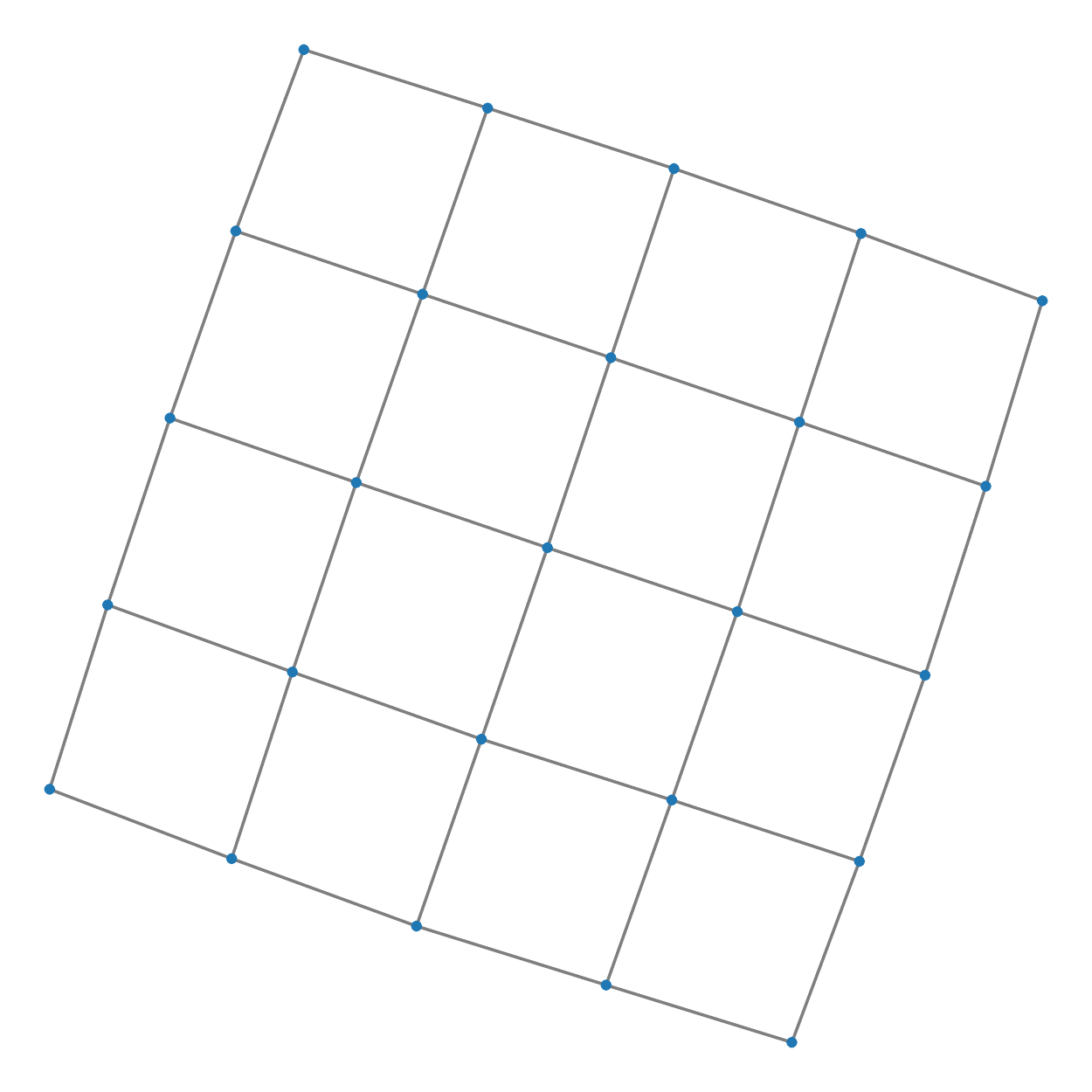}}&
  {\includegraphics[width=.24\columnwidth]{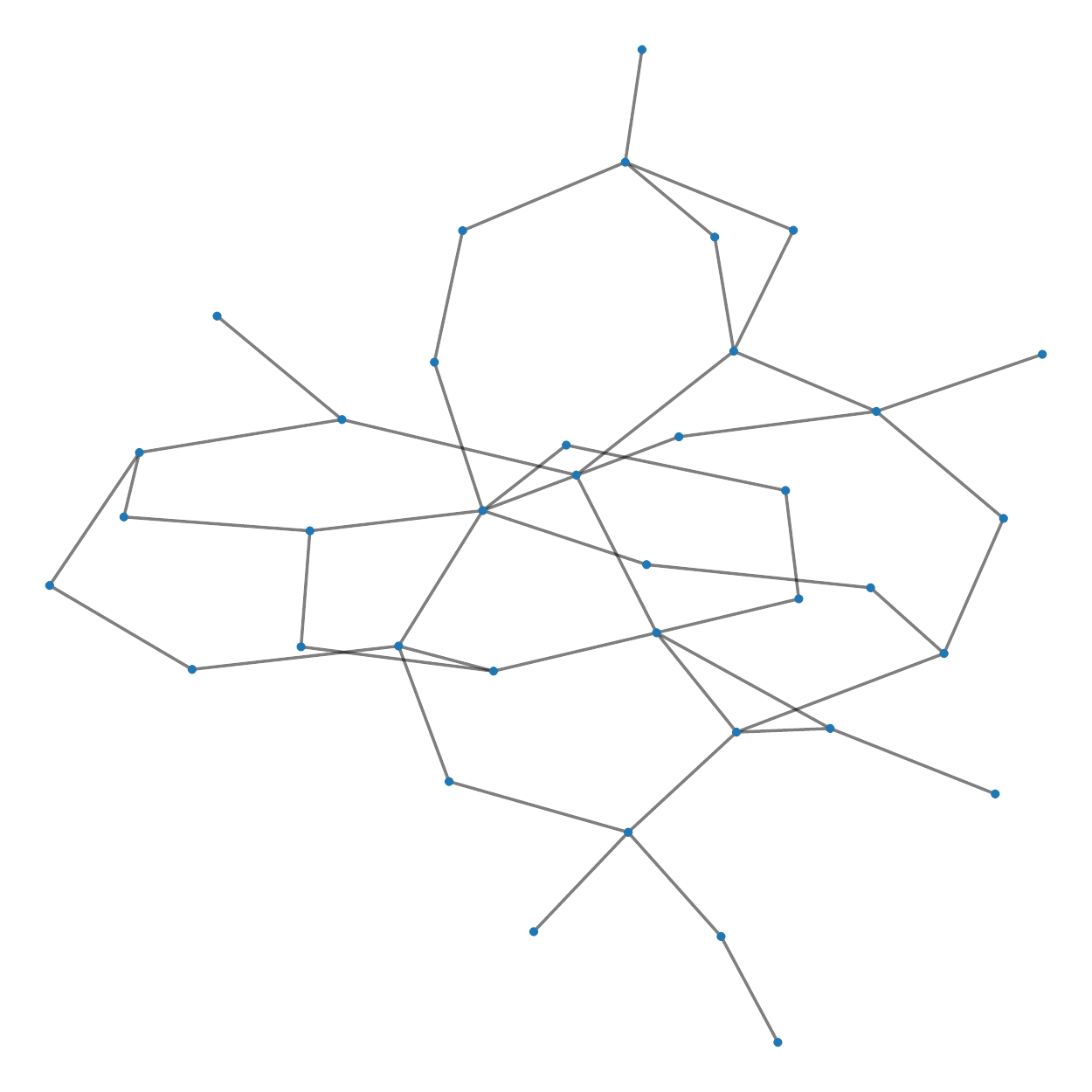}}&
  {\includegraphics[width=.24\columnwidth]{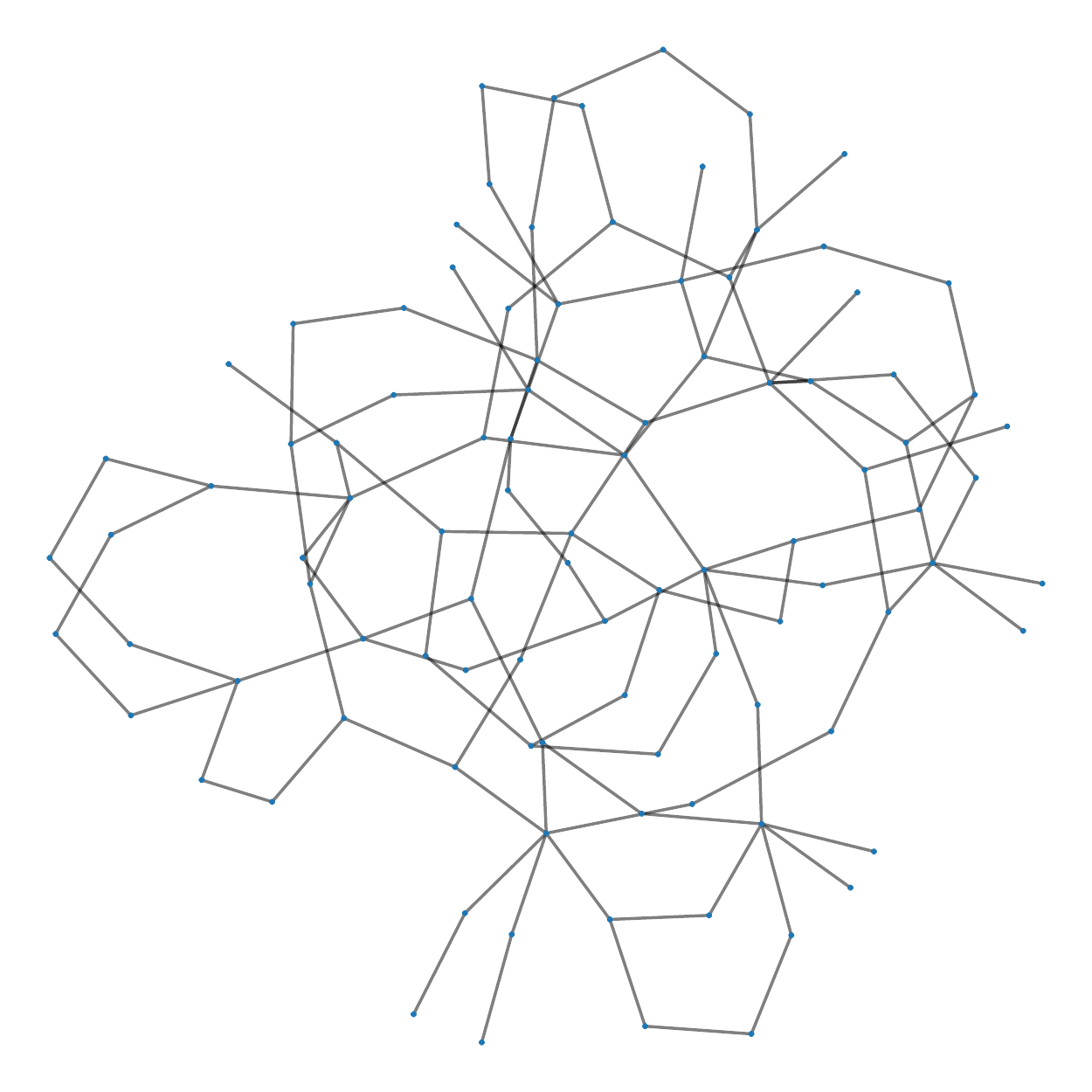}}\\
 \hline
\end{tabular}
}
\caption{Layout examples for \modelname, \modelnameStar,  $tsNET^{*}$, PivotMDS and \sgdgd.
}
\label{fig:visual_eval}
\end{figure}

Graph layout examples of \modelname are presented in \autoref{fig:visual_eval} alongside $tsNET^{*}$ and \sgdgd ones. \sgdgd drawings being all pleasing and only a few defects away from being perfect, we can use them to see how the layouts should look like. For \modelname, the dodecahedron and the grid graph structures can be observed but are severely distorted. It seems that \textit{topologically equivalent} nodes (\ie nodes that can be mapped to each other by an automorphism) are grouped together. Both drawings are therefore folded, which also emphasizes their symmetry. From what we experienced, this behavior might be caused by the \textit{compression} of the first stage training (see \autoref{sec:loss}) and too similar \textit{nodes features} and Chebyshev filters. Another explanation might be that the model had seen such small \textit{patterns} more often during the training stage and somehow overfitted on them.
On the two Rome graphs, the model has successfully laid the graphs structures out, but its tendency to group \textit{topologically equivalent} nodes leads to unbalanced edge lengths, edge crossings and overplots. On the other hand, \modelnameStar layouts are visually more pleasing. The dodecahedron structure can clearly be identified. Despite a lack of regularity, the grid layout is also acceptable. The two Rome graph layouts demonstrate \modelnameStar good performances. The model was able to separate topologically equivalent nodes, though they could have been repulsed a little more.
$tsNET^{*}$ also produces nodes overlaps where neighborhoods are similar, as it can be observed in the top right of its \textit{Rome graph 1} layout and on the right side of \textit{Rome graph 2}.

Overall \modelname and in particular \modelnameStar performed well even compared to OPT methods. The latter optimizing their cost function for a specific graph at a time, we could expect them to provide better results than a DL approach. Nevertheless, \modelname results acts as a proof-of-concept showing that we can learn unsupervised DL models to lay graphs out. It is therefore encouraging as we believe there is still a large room for improvement.  

\subsection{Limitations}
\label{sec:limitations}

The main limitation of \modelname is the technical need to set a maximum graph size \Nmax so that the architecture tensors size is static. Setting \Nmax to an arbitrarily high number might not be an option either since it would significantly increase the data pre-processing cost which is the most expensive step in \modelname. It is also not certain such model would learn if it is only fed with \textit{small} graphs (\ie with $N \ll N_{max}$), as each graph convolution kernel weight would be underfitted. 

Another limitation is the resources required to obtain a well-trained model. First, if the use of the trained model is straightforward, the model \textit{training} relies on many design choices that can only be efficiently made through a trials and errors process by an informed expert. 
Second, the computational resources required to train the model can be prohibitive. If Deep Learning-designed computers can easily handle small to mid-scale training (\eg \textit{finetuning}), heavier training (\eg \textit{pretraining}) can require to generate hundreds of thousands of graphs, which required the use of a Big Data platform in this experiment.

Finally, \modelname has not been tested on disconnected, weighted or directed graphs. Though the handling of these graph properties is straightforward with this framework, it is not part of the scope of this study.

\section{Conclusion}
\label{sec:conclusion}
We introduced \modelname, a Deep Learning based framework for graph drawing. \modelname proposes to adapt well-established Deep Neural Network architectures in image classification to compute the layout of an input graph by using Graph Convolutions. To the best of our knowledge, it is the first DL model trained to lay graphs out by directly optimizing a graph topology related cost function.

We provided an experimentation of the framework and compared its performances to graph drawing algorithms from the literature. The experiment showed that \modelname performs well compared to these algorithms despite some Deep Learning related limitations. The results highly suggest that Deep Learning is a promising approach for the future of graph drawing. It also implies that there exists a mathematical function that efficiently projects any graph structure into a drawing, and that can be learned by Deep Learning models.

Future work leads include trying out other \modelname implementations, meaning other Deep Neural Network architectures, loss functions (\eg \textit{stress}) and input node features (\eg \textit{node2vec}, \textit{DeepWalk}). Another interesting direction is to train and evaluate \modelname on other graph datasets or on specific graph families. Expanding the scale of the graphs size \modelname can handle, or apply it to other specific graph drawing applications are also promising leads for future work.

%\section*{Acknowledgments}
%This research project has been powered by Labo’s in the Sky with Data (LSD), the LaBRI data platform partially funded by the Region Nouvelle Aquitaine.

% \bibliographystyle{splncs03}
\bibliographystyle{splncs04}
\bibliography{on2dta}

\begin{thebibliography}{10}
\providecommand{\url}[1]{\texttt{#1}}
\providecommand{\urlprefix}{URL }
\providecommand{\doi}[1]{https://doi.org/#1}

\bibitem{gd2}
Ahmed, R., De~Luca, F., Devkota, S., Kobourov, S., Li, M.: {Graph Drawing via
  Gradient Descent, (GD)\textsuperscript{2}}. arXiv preprint arXiv:2008.05584
  (2020)

\bibitem{pivotmds}
{Brandes, Ulrik and Pich, Christian}: {Eigensolver methods for progressive
  multidimensional scaling of large data}. In: {International Symposium on
  Graph Drawing}. pp. 42--53. Springer (2006)

\bibitem{cohen2018approximating}
{Cohen-Steiner, David and Kong, Weihao and Sohler, Christian and Valiant,
  Gregory}: {Approximating the spectrum of a graph}. In: {Proceedings of the
  24th acm sigkdd international conference on knowledge discovery \& data
  mining}. pp. 1263--1271 (2018)

\bibitem{conover}
{Conover, William Jay and Iman, Ronald L}: {On multiple-comparisons
  procedures}. Tech. rep., {Technical report, Los Alamos Scientific Laboratory}
  (1979)

\bibitem{cnn4graph}
{Defferrard, Micha{\"e}l and Bresson, Xavier and Vandergheynst, Pierre}:
  {Convolutional neural networks on graphs with fast localized spectral
  filtering}. {arXiv preprint arXiv:1606.09375}  (2016)

\bibitem{deng2009imagenet}
{Deng, Jia and Dong, Wei and Socher, Richard and Li, Li-Jia and Li, Kai and
  Fei-Fei, Li}: {Imagenet: A large-scale hierarchical image database}. In:
  {2009 IEEE conference on computer vision and pattern recognition}. pp.
  248--255. Ieee (2009)

\bibitem{espadoto2020deep}
{Espadoto, Mateus and Hirata, Nina Sumiko Tomita and Telea, Alexandru C}: {Deep
  learning multidimensional projections}. {Information Visualization}
  \textbf{19}(3),  247--269 (2020)

\bibitem{frick1994gem}
{Frick, Arne and Ludwig, Andreas and Mehldau, Heiko}: {A fast adaptive layout
  algorithm for undirected graphs (extended abstract and system
  demonstration)}. In: {International Symposium on Graph Drawing}. pp.
  388--403. Springer (1994)

\bibitem{giovannangeli2020methodo}
{Giovannangeli, Loann and Bourqui, Romain and Giot, Romain and Auber, David}:
  {Toward automatic comparison of visualization techniques: Application to
  graph visualization}. {Visual Informatics}  \textbf{4}(2),  86--98 (2020)

\bibitem{grover2016node2vec}
{Grover, Aditya and Leskovec, Jure}: {node2vec: Scalable feature learning for
  networks}. In: {Proceedings of the 22nd ACM SIGKDD international conference
  on Knowledge discovery and data mining}. pp. 855--864 (2016)

\bibitem{haleem2019evaluating}
{Haleem, Hammad and Wang, Yong and Puri, Abishek and Wadhwa, Sahil and Qu,
  Huamin}: {Evaluating the readability of force directed graph layouts: A deep
  learning approach}. {IEEE computer graphics and applications}
  \textbf{39}(4),  40--53 (2019)

\bibitem{hammond2011cheby}
{Hammond, David K and Vandergheynst, Pierre and Gribonval, R{\'e}mi}: {Wavelets
  on graphs via spectral graph theory}. {Applied and Computational Harmonic
  Analysis}  \textbf{30}(2),  129--150 (2011)

\bibitem{sne}
{Hinton, Geoffrey and Roweis, Sam T}: {Stochastic neighbor embedding}. In:
  {NIPS}. vol.~15, pp. 833--840. Citeseer (2002)

\bibitem{he2016resnet}
{K. He and X. Zhang and S. Ren and J. Sun}: {Deep Residual Learning for Image
  Recognition}. In: {The IEEE Conference on Computer Vision and Pattern
  Recognition (CVPR)}. pp. 770--778 (June 2016)

\bibitem{kamada1989algorithm}
{Kamada, Tomihisa and Kawai, Satoru and others}: {An algorithm for drawing
  general undirected graphs}. {Information processing letters}  \textbf{31}(1),
   7--15 (1989)

\bibitem{kipf2016gcn}
{Kipf, Thomas N and Welling, Max}: {Semi-supervised classification with graph
  convolutional networks}. {arXiv preprint arXiv:1609.02907}  (2016)

\bibitem{kruiger2017tsnet}
{Kruiger, Johannes F and Rauber, Paulo E and Martins, Rafael M and Kerren,
  Andreas and Kobourov, Stephen and Telea, Alexandru C}: {Graph Layouts by
  t-SNE}. In: {Computer Graphics Forum}. vol.~36, pp. 283--294. {Wiley Online
  Library} (2017)

\bibitem{kruskalwallis}
{Kruskal, William H and Wallis, W Allen}: {Use of ranks in one-criterion
  variance analysis}. {Journal of the American statistical Association}
  \textbf{47}(260),  583--621 (1952)

\bibitem{kwanliuML4G}
{Kwon, Oh-Hyun and Crnovrsanin, Tarik and Ma, Kwan-Liu}: {What would a graph
  look like in this layout? a machine learning approach to large graph
  visualization}. {IEEE transactions on visualization and computer graphics}
  \textbf{24}(1),  478--488 (2017)

\bibitem{kwanliuDL4G}
{Kwon, Oh-Hyun and Ma, Kwan-Liu}: {A deep generative model for graph layout}.
  {IEEE Transactions on visualization and Computer Graphics}  \textbf{26}(1),
  665--675 (2019)

\bibitem{graphtsne}
{Leow, Yao Yang and Laurent, Thomas and Bresson, Xavier}: {GraphTSNE: a
  visualization technique for graph-structured data}. {arXiv preprint
  arXiv:1904.06915}  (2019)

\bibitem{deepwalk}
{Perozzi, Bryan and Al-Rfou, Rami and Skiena, Steven}: {Deepwalk: Online
  learning of social representations}. In: {Proceedings of the 20th ACM SIGKDD
  international conference on Knowledge discovery and data mining}. pp.
  701--710 (2014)

\bibitem{purchase1997aesthetic}
{Purchase, Helen}: {Which aesthetic has the greatest effect on human
  understanding?} In: {International Symposium on Graph Drawing}. pp. 248--261.
  Springer (1997)

\bibitem{purchase2002metrics}
{Purchase, Helen C}: {Metrics for graph drawing aesthetics}. {Journal of Visual
  Languages \& Computing}  \textbf{13}(5),  501--516 (2002)

\bibitem{purchase2012experimental}
{Purchase, Helen C}: {Experimental human-computer interaction: a practical
  guide with visual examples}. {Cambridge University Press} (2012)

\bibitem{purchase1995validating}
{Purchase, Helen C and Cohen, Robert F and James, Murray}: {Validating graph
  drawing aesthetics}. In: {International Symposium on Graph Drawing}. pp.
  435--446. Springer (1995)

\bibitem{Line}
{Tang, Jian and Qu, Meng and Wang, Mingzhe and Zhang, Ming and Yan, Jun and
  Mei, Qiaozhu}: {Line: Large-scale information network embedding}. In:
  {Proceedings of the 24th international conference on world wide web}. pp.
  1067--1077 (2015)

\bibitem{tsne}
{Van der Maaten, Laurens and Hinton, Geoffrey}: {Visualizing data using t-SNE.}
  {Journal of machine learning research}  \textbf{9}(11) (2008)

\bibitem{van2000mcl}
{Van Dongen, Stijn Marinus}: {Graph clustering by flow simulation}. Ph.D.
  thesis (2000)

\bibitem{ml4vis}
{Wang, Qianwen and Chen, Zhutian and Wang, Yong and Qu, Huamin}: {Applying
  Machine Learning Advances to Data Visualization: A Survey on ML4VIS}. {arXiv
  preprint arXiv:2012.00467}  (2020)

\bibitem{wang2019deepdrawing}
{Wang, Yong and Jin, Zhihua and Wang, Qianwen and Cui, Weiwei and Ma, Tengfei
  and Qu, Huamin}: {DeepDrawing: A deep learning approach to graph drawing}.
  {IEEE Transactions on Visualization and Computer Graphics}  \textbf{26}(1),
  676--686 (2019)

\bibitem{ambiguityvis}
{Wang, Yong and Shen, Qiaomu and Archambault, Daniel and Zhou, Zhiguang and
  Zhu, Min and Yang, Sixiao and Qu, Huamin}: {Ambiguityvis: Visualization of
  ambiguity in graph layouts}. {IEEE Transactions on Visualization and Computer
  Graphics}  \textbf{22}(1),  359--368 (2015)

\bibitem{ware2002cognitive}
{Ware, Colin and Purchase, Helen and Colpoys, Linda and McGill, Matthew}:
  {Cognitive measurements of graph aesthetics}. {Information visualization}
  \textbf{1}(2),  103--110 (2002)

\bibitem{ai4vis}
{Wu, Aoyu and Wang, Yun and Shu, Xinhuan and Moritz, Dominik and Cui, Weiwei
  and Zhang, Haidong and Zhang, Dongmei and Qu, Huamin}: {Survey on Artificial
  Intelligence Approaches for Visualization Data}. {arXiv preprint
  arXiv:2102.01330}  ({2021})

\bibitem{zheng2018sgdgd}
{Zheng, Jonathan X and Pawar, Samraat and Goodman, Dan FM}: {Graph drawing by
  stochastic gradient descent}. {IEEE transactions on visualization and
  computer graphics}  \textbf{25}(9),  2738--2748 (2018)

\end{thebibliography}

%%% APPENDIX

\clearpage
\appendix
\section*{Supplementary Materials}

This appendix presents supplementary materials and is organized as follows. 

\autoref{sec:appendix_loss} extensively presents the loss \modelname is trained to optimize.

\autoref{sec:appendix_trainInfos} presents the libraries used to implement \modelname, some specifications on their usage, and the hardware on which trainings and evaluations took place.

\autoref{sec:appendix_metrics} presents the intuitions and equations of the metrics used. \\

% \autoref{sec:appendix_stats} shows all the performances (average and standard deviation) measured during the experiment, as well as the \textit{p-values} of paiwise Conover-Iman post-hoc tests. It is organized following the order of benchmark in the paper: \autoref{sec:bench1} presents the results of the comparison of \modelname instances on various training methods (\autoref{sec:benchmark1}); \autoref{sec:bench2} presents the results of \textit{finetuned} \modelname variants comparisons with $tsNET$ and $tsNET^*$ (\autoref{sec:benchmark2}); and \autoref{sec:bench3} presents the results of the comparisons with state-of-the-art methods (\autoref{sec:benchmark3}). \\

To ease reading, we re-introduce the notations used both in the paper and in this supplementary material. Let $G(V,E)$ be a graph: $V$ is its set of nodes $\{v_i\}, i \in [1, N], N = |V|$
and $E \subseteq V \times V$ its set of edges. 
Graphs are considered simple and connected.
%\todo{j'ai déja vu des papiers qui ne le disaient pas clairement... je sais pas si c'est bien de pas le dire par contre}
$\delta$ is the distance matrix of $G$ and $\delta_{ij}$ is the shortest path length between nodes $v_i$ and $v_j$. 
Let nodes positions be encoded in a vector $X \in \mathds{R}^{N\times2}$ where $X_i$ is the 2D position of node $v_i$, $\left||X_i - X_j\right||$ relates to the euclidean distance between two points $X_i$ and $X_j$. \\

\section{Extended Loss Description}
\label{sec:appendix_loss}
The loss used to train \modelname is the adapted version of t-SNE to a graph context by Kruiger \etal~\cite{kruiger2017tsnet}. We would like to give a little more precision about the loss function and the intuition of the $C_{KL}$ term. All the following formula are taken from the Kruiger \etal~\cite{kruiger2017tsnet} paper \textit{Graph layout by t-SNE} and notations are adapted to our conventions. The loss was defined as: 

\begin{equation}
C = \lambda_{KL}C_{KL} + \frac{\lambda_c}{2N}\sum_i ||X_i||^2 - \frac{\lambda_r}{2N^2} \sum_{i,j \in V, i \neq j} \log\left(||X_i - X_j|| + \epsilon_r\right)
\end{equation}
where $C_{KL}$ is the main topology-related cost term (discussed below). The second term is a \textit{compression} that minimizes the scale of the drawing and is known to accelerate t-SNE convergence. The third term is a \textit{repulsion} that counter-balances the compression effects on the drawing. ($\lambda_{KL},\lambda_{c},\lambda_{r}$) are weights used to tune the loss function during the optimization. $\epsilon_r = \frac{1}{20}$ is a regularization constant.

The $C_{KL}$ (see \autoref{eq:CKL}) term is stated to be an adaptation to a graph context of the Kullback-Leibler divergence $D_{KL} = \sum\limits_i P(i) \log\frac{P(i)}{Q(i)}$ which measure the dissimilarity between two distributions of probabilities $P$ and $Q$. 
\begin{equation}
\label{eq:CKL}
C_{KL} = \sum_{i, j \in V, i \neq j} p_{ij} \log \frac{p_{ij}}{q_{ij}}
\end{equation}
where $p_{ij}$ is defined in \autoref{eq:P} and $q_{ij}$ is defined in \autoref{eq:Q}.

The intuition of $D_{KL}$ adaptation to a graph context is the following. For every pair of nodes $v_i, v_j \in V, i \neq j$, associate a probability $p_{ij}$ that the two nodes are neighbors. The probability associated to a pair of nodes depends on the distance between these nodes and their position in the graph (\eg dense or sparse regions). Thus, $p$ is defined as the distribution of probabilities that each pair of nodes is connected, computed from graph topology (see \autoref{eq:P}). $q$ is the distribution of probabilities that each pair of nodes is connected, computed from the nodes positions in the projected space of the layout (see \autoref{eq:Q}). Evaluating $D_{KL}$ with these two distributions of probabilities comes to study if the \textit{probabilities that all pairs of nodes are neighbors according to the graph topology} are similar to the \textit{probabilities that all pairs of nodes are neighbors according to their position in the layout}.

\begin{equation}
\label{eq:P}
p_{ij} = p_{ji} = \frac{p_{i|j} + p_{j|i}}{2N}
, p_{ii} = 0
\end{equation}
where $p_{i|j}$ is defined in \autoref{eq:pi_j}

\begin{equation}
\label{eq:pi_j}
p_{i|j} = \exp \left(-\frac{\delta_{ij}^2}{2\sigma_i^2}\right) / \sum_{k \in V, k \neq i} \exp \left(-\frac{\delta_{ik}^2}{2\sigma_i^2}\right)
, p_{i|i} = 0
\end{equation}
where $\sigma$ is a value associated to each node and depends on the node position in the graph. $\sigma$ can be find by binary search so that the \textit{perplexity} $\kappa_i = 2^{-\sum\limits_{j \in V} p_{i|j}\log_2p_{j|i}}$ for every node $v_i$ matches a defined value. In our experiment, the target \textit{perplexity} was dynamically set for each graph to depend on its number of nodes and ranges in [5; \Nmax/2], \ie [5; 64]. $\sigma$ typically gets low value in dense regions and high value in sparse regions.

\begin{equation}
\label{eq:Q}
q_{ij} = q_{ji} = \frac{(1+ ||X_i - X_j||^2)^{-1}}{\sum\limits_{k, l \in V \\ k \neq l}(1+ ||X_k - X_l||^2)^{-1}}
, q_{ii} = 0
\end{equation}

\section{Training informations}
\label{sec:appendix_trainInfos}
The experimentation were implemented with the Tensorflow-Keras API (version 2.4.1). Models were trained with Adam optimizer (with default parameters) for up to 200 epochs and an Early Stopping (with a patience of 30 for ptretraining, 20 for finetuning and from scratch training). The models have about one million of trainable parameters.

The models pretraining on Random graphs took place on a CPU cluster of 20 workers having 45GB of RAM each with a batch size of 400. ther trainings were run on a 8GB NVIDIA RTX 2080 Super GPU with a batch size of 32. Evaluations were processed on a dedicated computer with a Intel Xeon W-2123 3.60 GHz CPU with a 11GB NVIDIA GTX 1080 Ti GPU.

\section{Metrics Definitions}
\label{sec:appendix_metrics}
Here follows the formal definition of the aesthetic metrics used in the paper. Some were inverted so that all can be read as \textit{lower is better} (noted with *). See the paper for the formulas original paper. 

\textbf{Aspect Ratio*} is defined as the worst ratio between the drawing width and height after a serie of rotations $1 - \min_{\theta \in \{\frac{2\pi k}{N}, k \in [0, 1, ..., N-1]\}} \frac{\min(w_\theta, h_\theta)}{\max(w_\theta, h_\theta)}$\\

\textbf{Angular resolution*} is the ratio between the minimum angle formed by two edges on a node in the drawing and the optimal angle $\theta_G = \frac{2\pi}{d_{max}}$ that should be formed by the edges on the node with the maximum degree ($d_{max}$): $1 - \frac{\min_{(i,j), (j,k) \in E} \theta_{ijk}}{\theta_G}$. \\

\textbf{Edge crossings number} is the number of times edges cross each other in the drawing: $\sum\limits_{e1, e2 \in E, e1 \neq e2} \mathds{1}\{hasCrossing(e1, e2)\}$ where $hasCrossing$ is an elementary geometric function testing if two segments intersect. \\

\textbf{Cluster overlap} corresponds to the normalized sum of distances between nodes that are at a distance smaller than $r$ and not in the same cluster. The metric requires a neighborhood radius $r$ in the drawing and a clustering algorithm. Here, $r = 0.2$ and the clustering algorithm is MCL, an efficient deterministic clustering algorithm. The metric is formally defined as: $\sum\limits_{i \in V} \frac{\sum\limits_{u \in U_i}(1 - ||X_u - X_i||) *\mathds{1}\{MCL_i \neq MCL_u\}}{\sum\limits_{u \in U_i}(1 - ||X_u - X_i||)}$ where $U_i$ is the set of nodes that are at a distance less than $r$ from node $i$ and $MCL_i, i\in V$ is the cluster of node $v_i$ according to MCL algorithm.\\

\textbf{Neighborhood preservation*} is the sum, for each node, of the size of the intersection over union between its theoretical k-neighborhood $U$ and the set of its $|U|$ nearest nodes in the drawing called $Y$: $1 - \frac{1}{|V|} \sum\limits_{i \in V} \frac{|U_i \cap Y_i|}{|U_i \cup Y_i|}$. \\

\textbf{Stress.} $\sigma(X) = \frac{1}{N} \sum\limits_{i,j \in V} w_{ij}(||X_i - X_j|| - \delta_{ij})^2$, where $w_{ij}$ is set to $\delta_{ij}^{-2}$.\\

\textbf{Execution time}  is measured for every method in milliseconds ($ms$). For $(DNN)^2$, the graphs transformation in the required input formats (\textit{e.g.,} Chebyshev polynomials, nodes features) are taken into account.

% \clearpage
% \section{Statistic tables from the evaluations}
% \label{sec:appendix_stats}
% Each subsection corresponds to a benchmark in the paper (\ie \autoref{sec:benchmark1}, \ref{sec:benchmark2} and \ref{sec:benchmark3}). For each benchmark, the first table presents the performances of the compared methods; while the next tables present Conover test pairwise comparisons p-values.
% \input{appendix/stats.tex}

%%% END OF APPENDIX

\end{document}